\documentclass[sigconf]{acmart}

\usepackage{subfigure}

\AtBeginDocument{%
  \providecommand\BibTeX{{%
    \normalfont B\kern-0.5em{\scshape i\kern-0.25em b}\kern-0.8em\TeX}}}

\setcopyright{acmcopyright}
\copyrightyear{2018}
\acmYear{2018}
\acmDOI{XXXXXXX.XXXXXXX}

\acmConference[Conference acronym 'XX]{Make sure to enter the correct
  conference title from your rights confirmation emai}{June 03--05,
  2018}{Woodstock, NY}
\acmPrice{15.00}
\acmISBN{978-1-4503-XXXX-X/18/06}

\usepackage[justification=centering]{caption}
\usepackage[ruled,linesnumbered]{algorithm2e}
\usepackage{soul}

\begin{document}

\settopmatter{printacmref=false} 
\renewcommand\footnotetextcopyrightpermission[1]{} 
\pagestyle{empty} 

\title{Group-wise Reinforcement Feature Generation for Optimal and Explainable Representation Space  Reconstruction}

\author{Dongjie Wang}
\affiliation{%
	\institution{University of Central Florida}
	\state{FL}
	\country{USA}}
\email{wangdongjie@knights.ucf.edu}

\author{Yanjie Fu$^{{\dagger}}$}
\affiliation{%
	\institution{University of Central Florida}
	\state{FL}
	\country{USA}}
\email{yanjie.fu@ucf.edu}

\author{Kunpeng Liu}
\affiliation{%
	\institution{Portland State University}
	\state{OR}
	\country{USA}}
\email{kunpengliu0827@gmail.com}

\author{Xiaolin Li}
\affiliation{%
	\institution{Nanjing University}
	\state{Jiangsu}
	\country{China}}
\email{lixl@nju.edu.cn}

\author{Yan Solihin}
\affiliation{%
	\institution{University of Central Florida}
	\state{FL}
	\country{USA}}
\email{yan.solihin@ucf.edu}

\thanks{\small $^{\dagger}$Corresponding author}

\begin{abstract}

Representation (feature) space is an environment where data points are vectorized,  distances are computed, patterns are characterized, and geometric structures are embedded. Extracting a good representation space  is critical to address the curse of dimensionality, improve model generalization, overcome data sparsity, and increase the availability of classic models. 
Existing literature, such as feature engineering and representation learning, is limited in achieving full automation (e.g., over heavy reliance on intensive labor and empirical experiences), explainable explicitness (e.g., traceable reconstruction process and explainable new features), and flexible optimal (e.g., optimal feature space reconstruction is not embedded into downstream tasks).  
Can we simultaneously address the automation, explicitness, and optimal challenges in representation space reconstruction for a machine learning task?
To answer this question, we propose a  group-wise reinforcement generation perspective. 
We reformulate representation space reconstruction into an interactive process of nested feature generation and selection, where feature generation is to generate new meaningful and explicit features, and feature selection is to eliminate redundant features to control feature sizes. 
We develop a cascading reinforcement learning method that leverages three cascading Markov Decision Processes to learn optimal generation policies to automate the selection of features and operations and the feature crossing.
We design a group-wise generation strategy to cross a feature group, an operation, and another feature group to generate new features and find the strategy that can enhance exploration efficiency and augment reward signals of cascading agents.
Finally, we present extensive experiments to demonstrate the effectiveness, efficiency, traceability, and explicitness of our system.

\end{abstract}

\maketitle

\vspace{-0.1cm}
\section{Introduction}
Classic Machine Learning (ML) mainly includes data prepossessing, feature extraction, feature engineering, predictive modeling, and evaluation. 
The evolution of deep AI, however, has resulted in a new principled and widely used paradigm: i) collecting data, ii) computing data representations, and iii) applying ML models. 
Indeed, the success of ML algorithms highly depends on
data representation~\cite{bengio2013representation}. 
Building a good representation space is critical and fundamental because it can help to 1) identify and disentangle the underlying explanatory factors hidden in observed data, 2)  easy the extraction of useful information in predictive modeling,  3) reconstruct distance measures to form discriminative and machine-learnable patterns, 4) embed structure knowledge and priors into representation space and thus make classic ML algorithms available to complex graph, spatiotemporal, sequence, multi-media, or even hybrid data.

\begin{figure}[!htbp]
    \centering
    \includegraphics[width=1.0\linewidth]{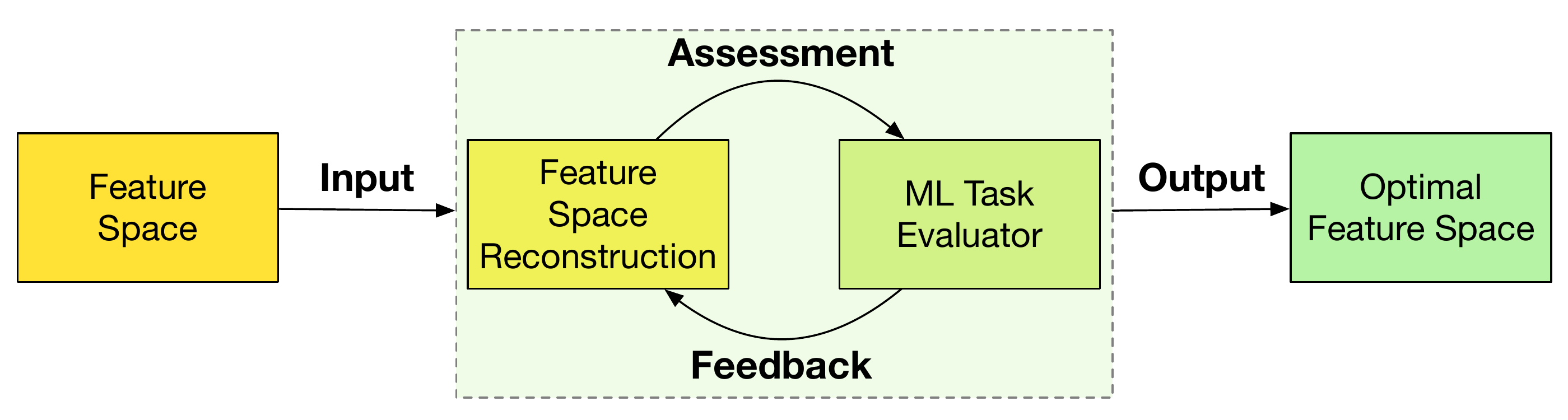}
    \vspace{-0.2cm}
    \captionsetup{justification=centering}
    \vspace{-0.5cm}
    \caption{We want to uncover the optimal feature space that is explainable and performs optimally in a downstream ML task by iteratively reconstructing the feature space.
    }
    \vspace{-0.25cm}
    \label{fig:intro_frame}
\end{figure}

In this paper, we study the problem of  learning to reconstruct an optimal and explainable feature representation space to advance a downstream ML task (\textbf{Figure \ref{fig:intro_frame}}).
Formally, given a set of original features, a prediction target, and a downstream ML task (e.g., classification, regression, ranking, detection), the objective is to automatically reconstruct an optimal and explainable set of features for the ML task. 

Prior literature has partially addressed the problem. 
The first relevant work is feature engineering, which designs preprocessing, feature extraction, selection~\cite{li2017feature,guyon2003introduction}, and generation~\cite{khurana2018feature} to extract a transformed representation of the data. These techniques are essential but labor-intensive, showing the low applicability of current ML practice in the automation of extracting a discriminative feature representation space.
\textbf{Issue 1 (full automation): } \emph{how can we make ML less dependent on feature engineering, construct ML systems faster, and expand the scope and applicability of ML?}
The second relevant work is representation learning, such as factorization~\cite{fusi2018probabilistic},  embedding~\cite{goyal2018graph}, and deep representation learning~\cite{wang2021reinforced,wang2021automated}. 
These studies are devoted to learning effective latent features. However, the learned features are implicit and non-explainable.
Such traits limit the deployment of these approaches in many application scenarios (e.g., patient and biomedical domains) that require not just high predictive accuracy but also trusted understanding and interpretation of underlying drivers. 
\textbf{Issue 2 (explainable explicitness): }
\emph{how can we assure that the reconstructing representation space is traceable and explainable?}
The third relevant work is learning based feature transformation, such as  principle component analysis~\cite{candes2011robust}, traversal transformation graph based feature generation~\cite{khurana2018feature}, sparsity regularization based feature selection~\cite{friedman2012fast,hastie2019statistical}.
These methods are either deeply embedded into or totally irrelevant to a specific ML model. 
For example, LASSO regression extracts an optimal feature subset for regression, but not for any given ML model. 
\textbf{Issue 3 (flexible optimal): } \emph{how can we create a framework to reconstruct a new representation space for any given predictor?} 
The three issues are well-known challenges. Our goal is to develop a new perspective to address these issues. 

\textbf{Our Contributions: A Traceable Group-wise Reinforcement Generation Perspective}.
We propose a novel principled framework to address the automation, explicitness, optimal issues in representation space reconstruction. 
We view feature generation and selection from the lens of Reinforcement Learning (RL).
We show that learning to reconstruct representation space can be accomplished by an interactive process of nested feature generation and selection, where feature generation is to generate new meaningful and explicit features, and feature selection is to remove redundant features to control feature sizes. 
We highlight that the human intuition and domain expertise in feature generation and selection can be formulated as machine-learnable policies. 
RL is an emerging technique to automatically generate experiences data and learn globally optimized policies.
Such traits have sparked considerable interest in recent years. 
We demonstrate that the iterative sequential feature generation and selection can be generalized as a RL task. 
We find that crossing features of high information distinctness is more likely to generate meaningful variables in a new representation space, and, thus, leveraging group-group crossing can accelerate the learning efficiency. 

\textbf{Summary of Proposed Approach}.
Based on our findings, we develop a generic and principled framework: group-wise reinforcement feature generation, for optimal and explainable representation space reconstruction.
This framework learns a representation space reconstructor that can
1)  \textbf{Goal 1: explainable explicitness:} provide traceable generation process and understand the meanings of each generated feature. 
2) \textbf{Goal 2: self optimization:} automatically generate an optimal feature set for a downstream ML task without much professional experience and human intervention; 
3)  \textbf{Goal 3: enhanced efficiency and reward augmentation:} 
enhance the generation and exploration speed in a large feature space and augment reward incentive signal to learn clear policies. 

To achieve Goal 1, we propose an iterative feature generation and selection  strategy, where the generation step is to apply a mathematical operation to two features to create a new feature and the selection step is to control the feature set size. 
This strategy allows us to explicitly trace the generation process and extract the semantic labels of generated features. 
To achieve Goal 2, we decompose feature generation into three Markov Decision Processes (MDPs): one is to select the first meta feature, one is to select an operation, and one is to select the second meta feature. We develop a new cascading agent structure to coordinate agents to share states and learn better selection policies for feature generation.
To achieve Goal 3, we design a group-operation-group based generation approach, instead of intuitive feature-operation-feature based generation, in order to accelerate representation space reconstruction. 
In particular, we first cluster the original features into different feature groups by maximizing intra-group feature cohesion and  inter-group feature distinctness, where we propose a novel feature-feature information distance. 
We then let agents select and cross two feature groups to generate multiple features each time. 
The benefits of this strategy are two folds: i) it explores feature space faster; ii) if we use feature-operation-feature based generation to add a single feature each time, the state of a feature set cannot be sufficiently changed, thus, restricting the agents from gaining enough reward to learn effective policies.
Instead, the group-operation-group based generation can alleviate this issue by augmenting the reward signal. 


\vspace{-0.1cm}
\section{Definitions and Problem Statement}

\subsection{Important Definitions}
\begin{definition}
\textbf{Feature Group.} 
We aim to reconstruct the feature space of such datasets $\mathcal{D}<\mathcal{F},y>$. 
Here, $\mathcal{F}$ is a feature set, in which each column denotes a feature and each row denotes a data sample;
$y$ is the target label set corresponding to samples.
To effectively and efficiently produce new features, we divide the feature set $\mathcal{F}$ into different feature groups via clustering, denoted by $\mathcal{C}$.
Each feature group  is a feature subset of  $\mathcal{F}$.

\end{definition}


\begin{definition}
\textbf{Operation Set.}
We perform a mathematical operation on existing features in order to generate new ones.
The collection of all operations is an operation set, denoted by $\mathcal{O}$.
There are two types of operations: unary and binary.
The unary operations include ``square'', ``exp'', ``log'', and etc.
The binary operations are ``plus'', ``multiply'', ``divide'', and etc.
\end{definition}

\begin{definition}
\textbf{Cascading Agent.}
To address the feature generation challenge, we develop a new cascading agent structure. 
This structure is made up of three agents: two feature group agents and one operation agent. Such cascading agents share state information and sequentially select feature groups and operations.
\end{definition}




\subsection{Problem Statement}
The research problem is learning to reconstruct an optimal and explainable feature representation space to advance a downstream ML task.
Formally, given a dataset $D<\mathcal{F},y>$ that includes an original feature set $\mathcal{F}$ and a target label set $y$, an operator set $\mathcal{O}$,   and a downstream ML task $A$ (e.g., classification, regression, ranking, detection), our objective is to automatically reconstruct an optimal and explainable feature set $\mathcal{F}^{*}$   that optimizes the performance indicator $V$ of the task $A$. 
The optimization objective  is to find a reconstructed feature set $\mathcal{\hat{F}}$ that maximizes:
\begin{equation}
\label{objective}
    \mathcal{F}^{*} = argmax_{\mathcal{\hat{F}}}( V_A(\mathcal{\hat{F}},y)),
\end{equation}
where $\mathcal{\hat{F}}$ can be viewed as a subset of a combination of  the original feature set $\mathcal{F}$ and the generated new features $\mathcal{F}^g$, and $\mathcal{F}^g$ is produced by applying the operations $\mathcal{O}$ to  the original feature set $\mathcal{F}$ via a certain algorithmic structure.

\section{Optimal and Explainable  Feature Space Reconstruction}
We present an overview, and then detail each technical component of our framework.

\begin{figure*}[!t]
    \centering
    \includegraphics[width=1.0\linewidth]{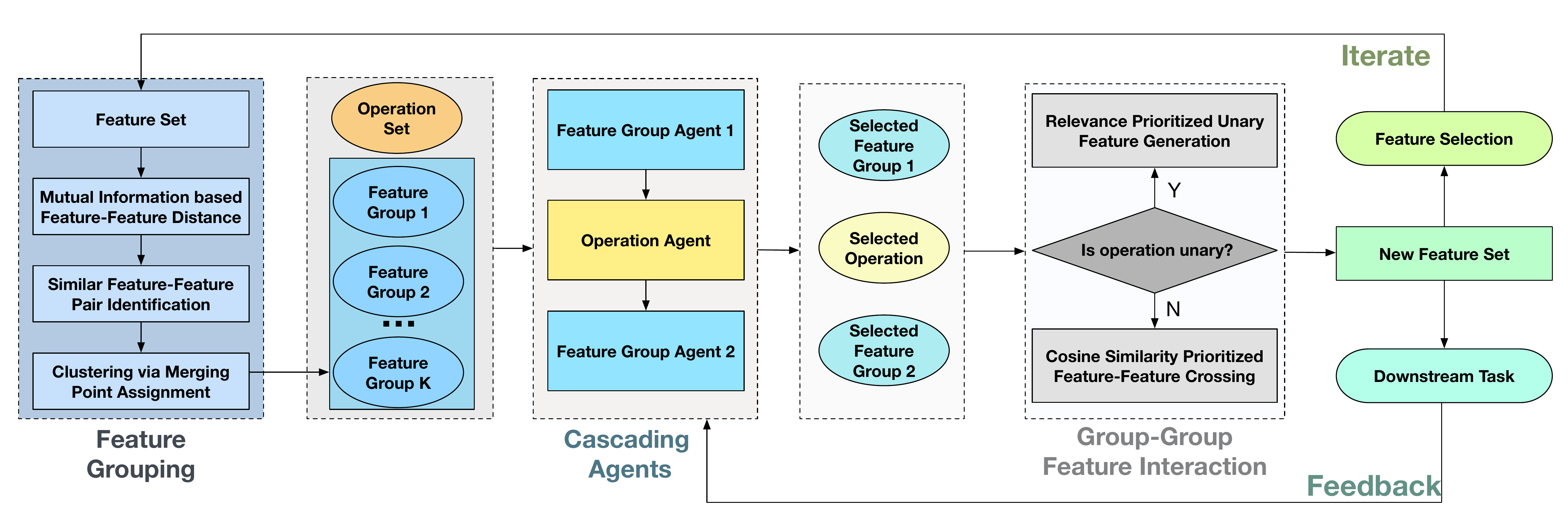}
    \vspace{-0.2cm}
    \captionsetup{justification=centering}
    \vspace{-0.33cm}
    \caption{An overview of GRFG. First, we cluster the feature set into feature groups. 
    Second, we employ cascading agents to select two feature groups and one operation.
    Next, we conduct group-group feature interaction to generate new features and combine them with original features to create a new feature set.
    Then, the updated feature set is fed into a downstream task to assess the selection process of cascading agents for parameter update.
    Meanwhile, we adopt feature selection to control the size of feature set and iterate the process until the best feature set is discovered or the maximum number of iterations reaches.
    }
    \vspace{-0.25cm}
    \label{fig:framework}
\end{figure*}

\subsection{Framework Overview}
Figure \ref{fig:framework} shows our proposed framework, \textbf{G}roup-wise \textbf{R}einforcement \textbf{F}eature \textbf{G}eneration (\textbf{GRFG}).
In the first step, we cluster the original features into different feature groups by maximizing intra-group feature cohesion and inter-group feature distinctness. 
In the second step, we leverage a group-operation-group strategy to cross two feature groups to generate multiple features each time. For this purpose, we develop a novel cascading reinforcement learning method to learn three agents to select the two most informative feature groups and the most appropriate operation from the operation set.
As a key enabling technique, the cascading reinforcement method will coordinate the three agents to share their perceived states in a cascading fashion, i.e., (agent1, state of the first feature group), (agent2, fused state of the operation and agent1's state), and (agent3, fused state of the second feature group and agent2's state), in order to learn better choice policies. 
After the two feature groups and operation are selected, we generate new features via a group-group crossing strategy. 
In particular, if the operation is unary, e.g., sqrt, we choose the feature group of higher relevance to target labels from the two feature groups, and apply the operation to the more relevant feature group  to generate new features. 
if the operation is binary,  we will choose the K most distinct feature-pairs from the two feature groups, and apply the binary operation to the chosen feature pairs to generate new features.
In the third step, we add the newly generated features to the original features to  create a generated feature set.
We feed the generated feature set into a downstream task to collect predictive accuracy as reward feedback to update  policy parameters.
Finally, we employ feature selection to eliminate redundant features and control the dimensonality of the newly generated feature set, which will be used as the original feature set to restart the iterations to regenerate new features until the maximum number of iterations is reached.

\noindent{\bf Comparison with prior literature.}
Automated feature engineering has recently attracted substantial research attention and has achieved great success. The transformation graph~\cite{khurana2018feature} and neural feature search~\cite{chen2019neural} are two typical methods that are closest to our task in existing literature. 
Instead of conducting personalized feature-feature crossing, the work in  ~\cite{khurana2018feature} generated new features by applying the selected operation to the entire feature set.
This generation strategy ignores features heterogeneity in a feature set, resulting in low-quality and sub-optimal features.
The study in ~\cite{chen2019neural} trained  a single recurrent neural network (RNN) for each feature to learn its feature transformation sequence for feature generation.
This strategy overlooks the feature distinctness in a feature set, restricting the method from producing  meaningful combined features.
Meanwhile, as the size of the feature set grows, the number of RNNs grows accordingly, which makes this work inefficient when dealing with large datasets.
To fill these gaps, our framework iteratively generates meaningful features via group-wise feature-feature interactions, which takes the feature heterogeneity into account. Moreover, we decompose the feature generation process into three MDPs and propose a simple cascading agent structure for it. Additionally,  our approach is a self-learning end-to-end  framework, which can be easily and flexibly applied to many scenarios.

\subsection{Generation-oriented Feature Clustering}
One of our key findings is that group-wise feature generation can accelerate the generation and exploration, and, moreover, augment reward feedback of agents to learn clear policies. 
Inspired by this finding, our system starts with generation oriented feature clustering, which aims to create feature groups that are meaningful for group-group crossing.  
Our another insight is that crossing features of high (low) information distinctness is more (less) likely to generate meaningful variables in a new representation space. 
As a result, unlike classic clustering, we aim to  cluster features into different feature groups, with the optimization objective of maximizing  inter-group feature information distinctness while minimizing intra-group feature information distinctness. 
To achieve this goal, we propose the \textbf{M-Clustering} for feature clustering, which starts with each feature as a feature group and then merges the closest feature group pair at each iteration. 

\textbf{Distance Between Feature Groups: A Group-level Relevance-Redundancy Ratio Perspective.} To achieve the goal of minimizing intra-group feature distinctness and maximizing inter-group feature distinctness,  we develop a new distance measure to quantify the distance between two feature groups. We highlight two interesting findings: 1) relevance to predictive target: if the relevance between one feature group and predictive target is similar to the relevance of another feature group and predictive target,  the two feature groups are similar; 2) mutual information: if the mutual information between the features of the two groups are large, the two feature groups are similar. 
Based on the two insights, we devise a feature group-group distance. 
The distance can be used to evaluate the distinctness of two feature groups, and, further, understand how likely crossing the two feature groups will generate more meaningful features. Formally, the distance is given by:
\begin{equation}
\vspace{-0.18cm}
    \label{fea_dis}
    dis(\mathcal{C}_i, \mathcal{C}_j) =
    \frac{1}{|\mathcal{C}_i|\cdot|\mathcal{C}_j|}
    \sum_{f_i\in \mathcal{C}_i}\sum_{f_j\in \mathcal{C}_j}\frac{|MI(f_i,y)-MI(f_j,y)|}{MI(f_i,f_j)+\epsilon},
\end{equation}
where $\mathcal{C}_i$ and $\mathcal{C}_j$ denote two feature groups, $|\mathcal{C}_i|$ and $|\mathcal{C}_j|$  respectively are the numbers of features in $\mathcal{C}_i$ and $\mathcal{C}_j$, $f_i$ and $f_j$ are two features in $\mathcal{C}_i$ and $\mathcal{C}_j$ respectively, $y$ is the target label vector.
In particular, $|MI(f_i,y)-MI(f_j,y)|$ quantifies the  difference in relevance  between $y$ and $f_i$, $f_j$.
If $|MI(f_i,y)-MI(f_j,y)|$ is small,   $f_i$ and $f_j$ have a more similar influence on classifying $y$;
$MI(f_i,f_j)+\epsilon$ quantifies the redundancy between $f_i$ and $f_j$.
$\epsilon$ is a small value that is used to prevent the denominator from being zero.
If $MI(f_i,f_j)+\epsilon$ is big, $f_i$ and $f_j$ share more information.
 
\textbf{Feature Group Distance based M-Clustering Algorithm:}
We develop a group-group distance instead of point-point distance,  and under such a group-level distance, the shape of the feature cluster could be non-spherical. Therefore, it is not appropriate to use K-means or density based methods. Inspired by agglomerative clustering,  given a feature set $\mathcal{F}$, we propose a three step method: 1) INITIALIZATION:  we regard each feature in $\mathcal{F}$ as a small feature cluster. 
2) REPEAT:  we calculate the information overlap between any two feature clusters and determine which cluster pair is the most closely overlapped.  We then merge two clusters into one and remove the two original clusters.
3) STOP CRITERIA:  we iterate the REPEAT step until the distance between the closest feature group pair reaches a certain threshold. Although classic stop criteria is to stop when there is only one cluster, using the distance between the closest feature groups as stop criteria can better help us to semantically understand, gauge, and identify the information distinctness among feature groups. It eases the implementation in practical deployment. 


\subsection{Cascading Reinforcement Feature Groups and Operation Selection}
\begin{figure}[!t]
    \centering
    \includegraphics[width=1.0\linewidth]{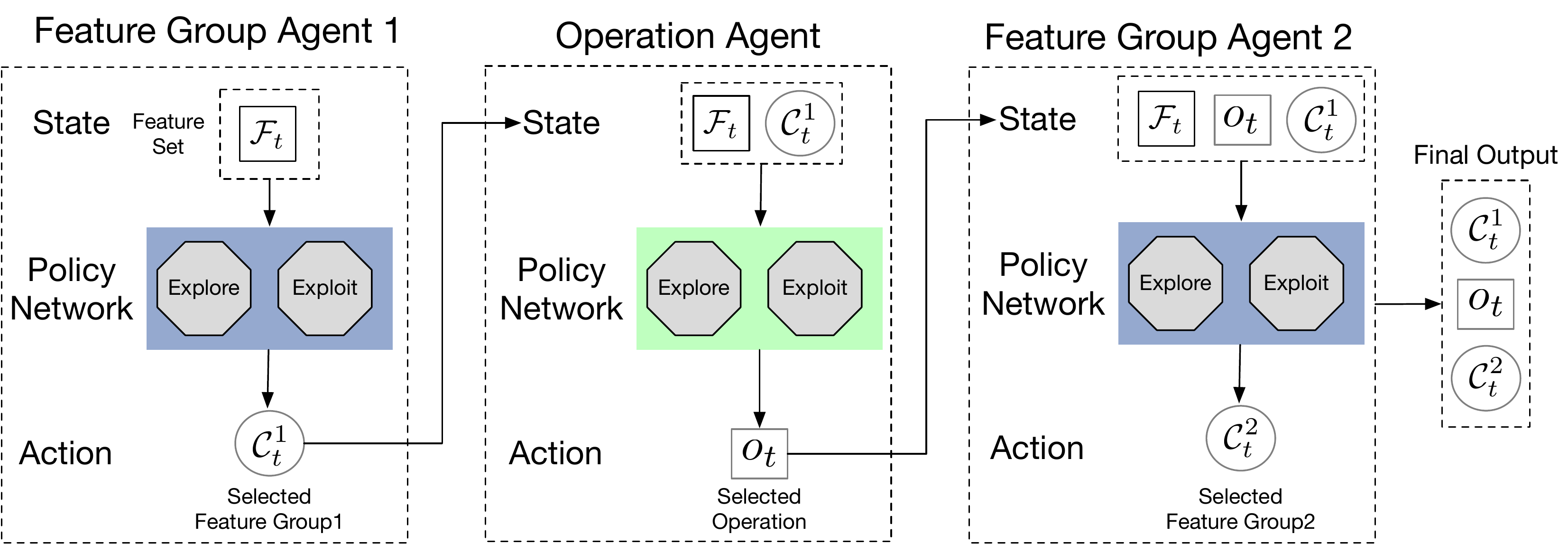}
    \vspace{-0.2cm}
    \captionsetup{justification=centering}
    \vspace{-0.2cm}
    \caption{The cascading agents are comprised of the feature group agent1, the operation agent, and the feature group agent2. They collaborate to choose two candidate feature groups and a single operation.}
    \vspace{-0.2cm}
    \label{fig:agents}
\end{figure}

To achieve group-wise feature generation, we need to  select a feature group, an operation, and another feature group to perform group-operation-group based crossing. Two key findings inspire us to leverage cascading reinforcement. 
\textbf{Firstly}, we highlight that although it is hard to define and program the optimal selection criteria of feature groups and operation, we can view selection criteria as a form of machine-learnable policies. 
We propose three agents to learn the policies by trials and errors. 
\textbf{Secondly}, we find that the three selection agents are cascading in a sequential auto-correlated decision structure, not independent and separated. 
Here, ``cascading'' refers to the fact that within each iteration agents  make decision sequentially, and downstream agents await for the completion of an upstream agent.  The decision of an upstream agent will change the environment states of downstream agents.
As shown in Figure ~\ref{fig:agents}, the first agent picks the first feature group based on the state of the entire feature space, the second agent picks the operation based on the entire feature space and the selection of the first agent, and the third agent chooses the second feature group based on the entire feature space and the selections of the first and second agents. 

We next propose two generic metrics to quantify the usefulness (reward) of  a feature set, and then form three MDPs to learn three selection policies. 

\smallskip
\noindent\textbf{Two Utility Metrics for Reward Quantification.} The two utility metrics are from  the supervised and unsupervised perspectives. 

\noindent\ul{Metric 1: Integrated Feature Set Redundancy and Relevance.}  
We propose a metric to quantify feature set utility from an information theory perspective: a higher-utility feature set has less redundant information and is more relevant to prediction targets.
Formally, given a feature set $\mathcal{F}$ and a predictive target label $y$, such utility metric can be calculated by 
\begin{equation}
    U(\mathcal{F}|y) = -\frac{1}{|\mathcal{F}|^2}\sum_{f_i, f_j \in \mathcal{F}} MI(f_i, f_j) + \frac{1}{|\mathcal{F}|}\sum_{f\in \mathcal{F}}MI(f,y),
\end{equation}
where $MI$ is the mutual information, $f_i, f_j, f$ are features in $\mathcal{F}$ and $|\mathcal{F}|$ is the size of the feature set $\mathcal{F}$.

\noindent\ul{Metric 2: Downstream Task Performance.} 
Another utility metric is whether this feature set can improve a downstream task (e.g., regression, classification). We use a downstream predictive task performance indicator (e.g., 1-RAE, Precision, Recall, F1) as a utility metric of a feature set. 


\smallskip
\noindent\textbf{Learning Selection Agents of Feature Group 1, Operation, and Feature Group 2.} 
Leveraging the two metrics, we next develop three MDPs to learn three agent policies to select the best feature group 1, operation, feature group 2. 

\noindent\ul{\textit{Learning the Selection Agent of Feature Group 1.}} 
The feature group agent 1 iteratively select the best meta feature group 1.
Its learning system includes: \textbf{i) Action:} its action 
at the $t$-th iteration is the meta feature group 1  selected from the feature groups of the previous iteration, denoted group $a_t^1 = \mathcal{C}^1_{t-1}$.
\textbf{ii) State:} its state at the $t$-th iteration is a vectorized representation of the generated feature set of the previous iteration.
Let $Rep$ be a state representation method, the state can be denoted by $s_t^1 = Rep(\mathcal{F}_{t-1})$.
We will discuss the state representation method in the next section.
\textbf{iii) Reward:} its reward at the $t$-th iteration is the utility score the selected feature group 1, denoted by  $\mathcal{R}(s_t^1,a_t^1)=U(\mathcal{F}_{t-1}|y)$.


\noindent\ul{\textit{Learning the Selection Agent of Operation.}} 
The operation agent will iteratively select the best operation (\textit{e.g.} +, -) from an operation set as a feature crossing tool for feature generation.
Its learning system includes: \textbf{i) Action:} its action at the $t$-th iteration is the selected operation, denoted by $a_t^o = o_t$.
\textbf{ii) State:} its state at the $t$-th iteration is the combination of $Rep(\mathcal{F}_{t-1})$ and the representation of the  feature group selected by the previous agent, denoted by $s^o_t = Rep(\mathcal{F}_{t-1}) \oplus Rep(\mathcal{C}_{t-1}^1)$, where $\oplus$ indicates the concatenation operation. 
\textbf{iii) Reward:} The selected operation will be used to generate new features by feature crossing. We combine such new features with the original feature set to form a new feature set.
Thus, the feature set at the $t$-th iteration is $\mathcal{F}_t = \mathcal{F}_{t-1} \cup g_t$, where $g_t$ is the generated new features.
The reward of this iteration is the improvement in the utility score of the feature set compared with the previous iteration, denoted by $\mathcal{R}(s_t^o,a_t^o) = U(\mathcal{F}_t|y) - U(\mathcal{F}_{t-1}|y)$.


\noindent\ul{\textit{Learning the Selection Agent of Feature Group 2.}} 
The feature group agent 2 will iteratively select the best meta feature group 2 for feature generation.
Its learning system includes: \textbf{i) Action:} its action at the $t$-th iteration is the meta feature group 2 selected from the clustered feature groups of the previous iteration, denoted by $a_t^2 = \mathcal{C}^2_t$.
\textbf{ii) State:} its state at the $t$-th iteration is combination of $Rep(\mathcal{F}_{t-1})$, $Rep(\mathcal{C}_{t-1}^1)$ and the vectorized representation of the  operation selected by the operation agent, denoted by $s_t^2 = Rep(\mathcal{F}_{t-1})\oplus Rep(\mathcal{C}_{t-1}) \oplus Rep(o_t)$.
\textbf{iii) Reward:} its reward at the $t$-th iteration is improvement of the feature set utility and the feedback of the downstream task, denoted by $\mathcal{R}(s_t^2, a_t^2) = U(\mathcal{F}_t|y)-U(\mathcal{F}_{t-1}|y)+V_{A_t}$, where $V_A$ is the performance (e.g., F1) of a downstream predictive task.

\begin{figure}[t]
    \centering
    \includegraphics[width=1.0\linewidth]{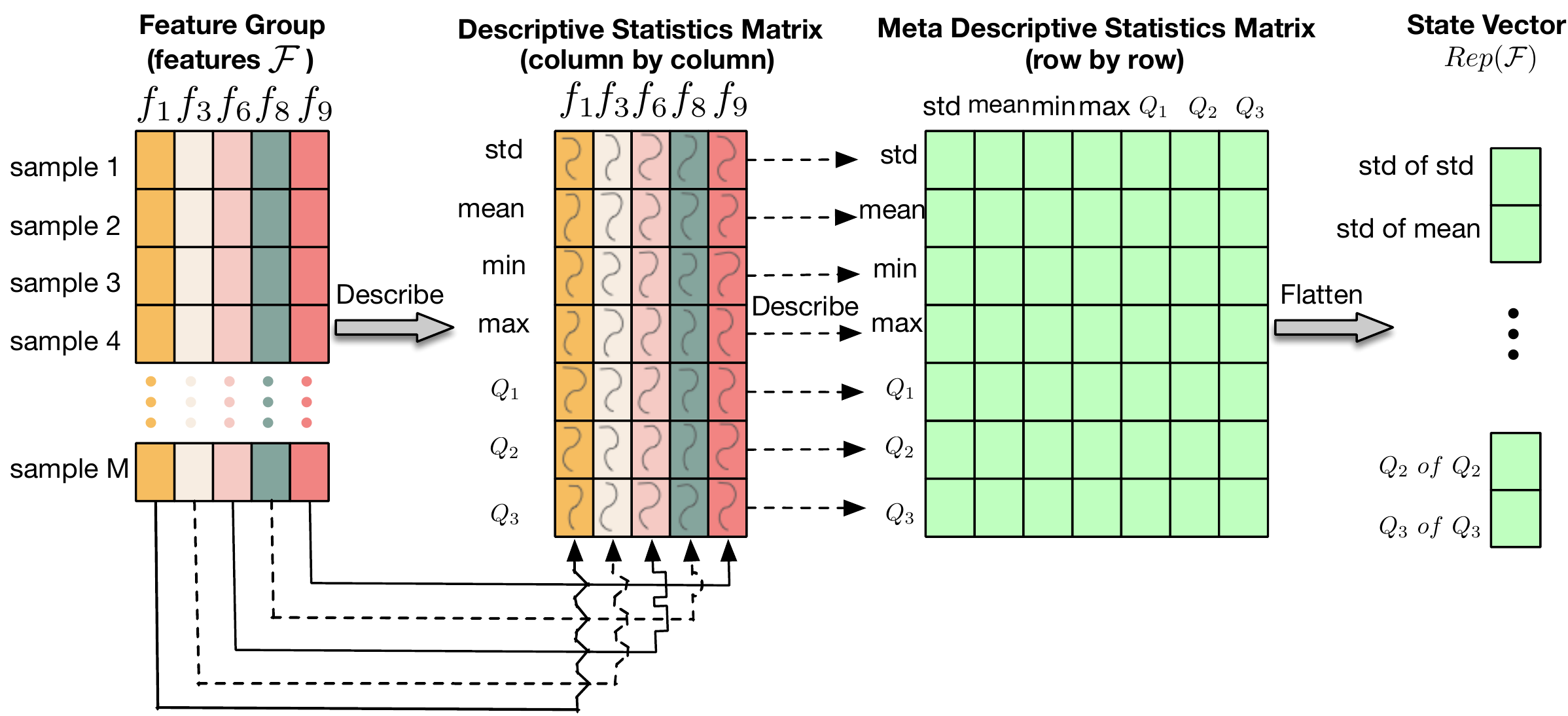}
    \vspace{-0.3cm}
    \captionsetup{justification=centering}
    \vspace{-0.2cm}
    \caption{State Representation. Given a feature group $\mathcal{F}$ consisting of several features, we calculate the descriptive statistics of $\mathcal{F}$ column-by-column, then row-by-row to get a meta descriptive statistics matrix.
    Then, we flat the matrix to obtain the state representation vector $Rep(\mathcal{F})$.}
    \vspace{-0.2cm}
    \label{fig:state_repr}
\end{figure}

\smallskip
\noindent\textbf{State Representation of a Feature Group and an Operation.}
We propose to map a feature group to a vector that characterizes the \textbf{State} of the given feature group. 
In detail, given a feature group $\mathcal{F}$, we first calculate the descriptive statistics (\textit{i.e.} count, standard deviation, minimum, maximum, first , second , and third quartile) column by column.
Then, row by row, we calculate the descriptive statistics of the outcome of the previous step to obtain the descriptive matrix that shape is $\mathbb{R}^{7\times 7}$.
After that, we obtain the feature feature's representation $Rep(\mathcal{F})\in \mathbb{R}^{1\times 49}$ by flatting the descriptive matrix.
A fixed-size state vector is produced by the representation method, which accommodates the varying size of the feature set at each  generation iteration.
Second, for the operation, we use the one-hot encoding as its representation $Rep(o)$.

\smallskip
\noindent\textbf{Solving the Optimization Problem.} 
We train the three agents by maximizing the discounted and cumulative reward during the iterative feature generation process.
In other words, we encourage the cascading agents to collaborate to generate a feature set that is independent, informative, and performs well in the downstream task.
To accomplish this goal, we minimize the temporal difference error $\mathcal{L}$ converted from the Bellman equation, given by:
\begin{equation}
\label{Q_update}
    \mathcal{L} = Q(s_{t},a_t) - (\mathcal{R}(s_t, a_t) + \gamma * \text{max}_{a_{t+1}}Q(s_{t+1},a_{t+1})),
\end{equation}
where $\gamma \in [0\sim 1]$ is the discounted factor; $Q$ denotes the $Q$ function estimated by deep neural networks. After agents converge, we expect to discover the optimal policy $\pi^*$ that can choose the most appropriate action (\textit{i.e.} feature group or operation) based on the state via the Q-value, which can be formulated as follows:
\begin{equation}
    \label{rl_policy}
    \pi^*(a_t|s_t) = \text{argmax}_a Q(s_t,a).
\end{equation}

\subsection{Group-wise Feature Generation}
\label{group_generation}
We found that using group-level crossing can generate more features each time, and, thus, accelerate exploration speed, augment reward feedback by adding significant amount of features, and effectively learn policies.  
The selection results of our reinforcement learning system include \textbf{two generation scenarios}: (1) selected are a binary operation and two feature groups;  (2) selected are a unary operation and two feature groups. 
However, a challenge arises: what are the most effective generation strategy for the two scenarios?
We next propose two  strategies for the two scenarios.

\textbf{Scenario 1: Cosine Similarity Prioritized Feature-Feature Crossing.} 
We highlight that it is more likely to generate an informative features by crossing two features that are less overlapped in terms of information. 
We propose a simple yet efficient strategy, that is, to select the top K dissimilar feature pairs between two feature groups. 
Specifically, we first cross two selected feature groups to prepare feature pairs. We then compute the cosine similarities of all feature pairs. Later, we rank and select the top K dissimilar feature pairs. Finally, we apply the operation to the top K selected feature pairs to generate K new features.

\textbf{Scenario 2: Relevance Prioritized  Unary Feature Generation. }
When selected  are an unary operation and two feature groups, we directly apply the operation to the feature group that is more relevant to target labels. Given a feature group $C$, we use the average mutual information between all the features in $C$ and the prediction target $y$ to quantify the relevance between the feature group and the prediction targets, which is given by: 
$
   rel =  \frac{1}{|\mathcal{C}|}\sum_{f_i\in \mathcal{C}} MI(f_i,y) 
$, 
where $MI$ is a function of mutual information. 
After the more relevant feature group is identified, we apply the unary operation to each feature of the feature group to generate new features. 

\textbf{Post-generation Processing.} 
After feature generation, we combine the newly generated features with  the original feature set to form an updated feature set, which will be fed into a downstream task to evaluate predictive performance.
Such performance is exploited as reward feedback to update the policies of the three cascading agents in order to optimize the next round of feature generation.
To prevent feature number explosion during the iterative generation process, 
we use a feature selection step to control feature size. 
When the size of the new feature set exceeds a feature size tolerance threshold, we leverage the K-best feature selection method to reduce the feature size. Otherwise, we don't perform feature selection. 
We use the tailored new feature set as the original feature set of the next iteration. 

Finally, when the maximum number of iterations is reached, the algorithm  returns the optimal feature set $\mathcal{F^{*}}$ that has the best downstream performance over the entire exploration.

\vspace{-0.1cm}

\begin{table*}[!htbp]
\centering
\vspace{-0.2cm}
\caption{Overall performance comparison. `C' for classification and `R' for regression.}
\vspace{-0.3cm}
\label{table_overall_perf}
\setlength{\tabcolsep}{2.5mm}{
\begin{tabular}{c|c|c|c|c|c|c|c|c|c|c|c}
\hline
Dataset            & Source   & C/R & Samples & Features & RDG  & ERG & LDA & AFT   & NFS   & TTG  & GRFG           \\ \hline
Higgs Boson        & UCIrvine & C  & 50000   & 28       & 0.683 & 0.674 & 0.509 & 0.711 & 0.715 & 0.705 & \textbf{0.719} \\ \hline
Amazon Employee    & Kaggle   & C   & 32769   & 9        & 0.744 & 0.740 & 0.920  & 0.943 & 0.935 & 0.806 & \textbf{0.946} \\ \hline
PimaIndian         & UCIrvine & C   & 768     & 8        & 0.693 & 0.703 & 0.676  & 0.736 & 0.762 & 0.747 & \textbf{0.776} \\ \hline
SpectF             & UCIrvine & C   & 267     & 44       & 0.790 & 0.748 & 0.774  & 0.775 & 0.876 & 0.788 & \textbf{0.878} \\ \hline
SVMGuide3  & LibSVM & C & 1243 & 21 & 0.703 & 0.747 & 0.683 & 0.829 & 0.831 & 0.766 & \textbf{0.850} \\ \hline
German Credit      & UCIrvine & C   & 1001    & 24       & 0.695 & 0.661 & 0.627  & 0.751 & 0.765 & 0.731 & \textbf{0.772} \\ \hline
Credit Default     & UCIrvine & C   & 30000   & 25       & 0.798 & 0.752 & 0.744  & 0.799 & 0.799 & \textbf{0.809} & 0.800 \\ \hline
Messidor\_features & UCIrvine & C   & 1150    & 19       & 0.673 & 0.635 & 0.580  & 0.678 & 0.746 & 0.726 & \textbf{0.757} \\ \hline
Wine Quality Red   & UCIrvine & C   & 999     & 12       & 0.599 & 0.611 & 0.600  & 0.658 & 0.666 & 0.647 & \textbf{0.686} \\ \hline
Wine Quality White & UCIrvine & C   & 4900    & 12       & 0.552 & 0.587 & 0.571  & 0.673 & 0.679 & 0.638 & \textbf{0.685} \\ \hline
SpamBase           & UCIrvine & C   & 4601    & 57       & 0.951 & 0.931 & 0.908  & 0.951 & 0.955 & \textbf{0.961} & 0.958 \\ \hline
AP-omentum-ovary            & OpenML & C   & 275    & 10936        & 0.711 & 0.705 & 0.117  & 0.783 & 0.804 & 0.795 & \textbf{0.818} \\ \hline
Lymphography       & UCIrvine & C   & 148     & 18       & 0.654 & 0.638 & 0.737 & 0.833 & 0.859 & 0.846 & \textbf{0.866} \\ \hline
Ionosphere         & UCIrvine & C   & 351     & 34       & 0.919 & 0.926 & 0.730  & 0.827 & 0.949 & 0.938 & \textbf{0.960} \\ \hline
Bikeshare DC       & Kaggle   & R   & 10886   & 11       & 0.483 & 0.571 & 0.494  & 0.670 & 0.675 & 0.659 & \textbf{0.681} \\ \hline
Housing Boston     & UCIrvine & R   & 506     & 13       & 0.605 & 0.617 & 0.174 & 0.641 & 0.665 & 0.658 & \textbf{0.684} \\ \hline
Airfoil            & UCIrvine & R   & 1503    & 5        & 0.737 & 0.732 & 0.463  & 0.774 & 0.771 & 0.783 & \textbf{0.797} \\ \hline
Openml\_618        & OpenML   & R   & 1000    & 50       & 0.415 & 0.427  & 0.372 & 0.665 & 0.640 & 0.587 & \textbf{0.672} \\ \hline
Openml\_589        & OpenML   & R   & 1000    & 25       & 0.638 & 0.560 & 0.331  & 0.672 & 0.711 & 0.682 & \textbf{0.753} \\ \hline
Openml\_616        & OpenML   & R   & 500     & 50       & 0.448 & 0.372 & 0.385  & 0.585 & 0.593 & 0.559 & \textbf{0.603} \\ \hline
Openml\_607        & OpenML   & R   & 1000    & 50       & 0.579 & 0.406 & 0.376  & 0.658 & 0.675 & 0.639 & \textbf{0.680} \\ \hline
Openml\_620        & OpenML   & R   & 1000    & 25       & 0.575 & 0.584 & 0.425 & 0.663 & 0.698 & 0.656 & \textbf{0.714} \\ \hline
Openml\_637        & OpenML   & R   & 500     & 50       & 0.561 & 0.497 & 0.494  & 0.564 & 0.581 & 0.575 & \textbf{0.589} \\ \hline
Openml\_586        & OpenML   & R   & 1000    & 25       & 0.595 & 0.546 & 0.472  & 0.687 & 0.748 & 0.704 & \textbf{0.783} \\ \hline
\end{tabular}}
\vspace{-0.2cm}
\end{table*}

\section{Experiments}
 

\subsection{Experimental Setup}
\subsubsection{Data Description}
We used 24 publicly available datasets from UCI~\cite{uci}, 
 LibSVM ~\cite{libsvm},
 Kaggle ~\cite{kaggle}, and OpenML ~\cite{openml} to conduct experiments.
The 24 datasets involves 14 classification tasks and 10 regression tasks.
Table \ref{table_overall_perf} shows the statistics of the data. 

\subsubsection{Evaluation Metrics}
We used the F1-score to evaluate the recall and precision of classification tasks. 
We used  1-relative absolute error (RAE) to evaluate the accuracy of regression tasks. 
Specifically, $\text{1-RAE} = 1 - \frac{\sum_{i=1}^{n} |y_i-\check{y}_i|}{\sum_{i=1}^{n}|y_i-\bar{y}_i|}$, where $n$ is the number of data points, $y_i, \check{y}_i, \bar{y}_i$ respectively denote golden standard  target values, predicted target values, and the mean of golden standard targets. 

\subsubsection{Baseline Algorithms}
\label{baseline}
We compared our method with five widely-used feature generation methods: 
(1) \textbf{RDG} randomly selects feature-operation-feature pairs for feature generation; 
(2) \textbf{ERT} is a expansion-reduction method, that applies operations to each feature to expand the feature space and selects significant features from the larger space as new features. 
(3) \textbf{LDA}~\cite{blei2003latent} extracts latent features via matrix factorization.
(4) \textbf{AFT}~\cite{horn2019autofeat}  is an enhanced ERT implementation that iteratively explores feature space and adopts a multi-step feature selection relying on L1-regularized linear models.
(5) \textbf{NFS}~\cite{chen2019neural} mimics feature transformation trajectory for each feature and optimizes the entire feature generation process through reinforcement learning.
(6) \textbf{TTG} ~\cite{khurana2018feature} records the feature generation process using a transformation graph, then uses reinforcement learning to explore the graph to determine the best feature set.

Besides, we developed four variants of GRFG to validate the impact of each technical component: 
(i) $\textbf{GRFG}^{-c}$ removes the clustering step of GRFG and generate features by feature-operation-feature based crossing, instead of group-operation-group based crossing. 
(ii) $\textbf{GRFG}^{-d}$ utilizes the euclidean distance as the measurement of M-Clustering.
(iii) $\textbf{GRFG}^{-u}$ selects a feature group at random from the feature group set, when the  operation is unary.
(iv) $\textbf{GRFG}^{-b}$ randomly selects features from the larger feature group to align two feature groups when the operation is binary. 
We adopted random forest, a robust ensemble predictor, as the downstream ML model, in order to ensure the changes of results are mainly caused by the feature space reconstruction, not randomness or variance of the predictor.
We performed 5-fold stratified cross-validation in all  experiments, instead of a simple 70\%-30\% split.

\subsubsection{Hyperparameters, Source Code and Reproducibility}
The operation set consists of \textit{square root, square, cosine, sine, tangent, exp, cube, log, reciprocal, sigmoid, plus, subtract, multiply, divide}.
We limited iterations (epochs) to 30, with each iteration consisting of 15 exploration steps.
When the number of generated features is twice of the original feature set size, we performed feature selection to control feature size.
In GRFG, all agents were constructed using a DQN network with two linear layers activated by the RELU function.
We optimized DQN using the Adam optimizer with a 0.01 learning rate, and set the limit of the experience replay memory as 32 and the batch size as 8.
The parameters of all the baseline models are defined based on the recommendations of corresponding papers.
For other detailed experimental settings, please check the code released in the Abstract section.

\subsubsection{Environmental Settings}
All experiments were conducted on the Ubuntu 18.04.5 LTS operating system, Intel(R) Core(TM) i9-10900X CPU@ 3.70GHz, and 1 way SLI RTX 3090 and 128GB of RAM, with the framework of Python 3.8.5 and PyTorch 1.8.1.

\vspace{-0.1cm}
\subsection{Experimental Results}

\subsubsection{Overall Comparison}
This experiment aims to answer:
\textit{Can our method effectively construct quality feature space and improve a downstream task?} 
Table \ref{table_overall_perf} shows the comparison of our method with six baseline models on the 24 datasets in terms of F1 score or 1-RAE.
We observed that GRFG ranks first on most datasets and ranks second on ``Credit Default'' and ``SpamBase''.
The underlying driver is that the personalized feature crossing strategy in GRFG considers feature-feature distinctions when generating new features.
Besides, the observation that GRFG outperforms random-based  (RDG) and expansion-reduction-based (ERG, AFT) methods 
shows that the agents can share states and rewards in a cascading fashion, and, thus learn an effective policy to select optimal crossing features and operations.
Moreover, because our method is a self-learning end-to-end framework, users can treat it as a tool and easily apply it to different datasets regardless of implementation details.
Thus, compared with state-of-the-art baselines (NFS, TTG), our method is more practical and automated in real application  scenarios.

\begin{figure*}[htbp]
\centering
\subfigure[PimaIndian]{
\includegraphics[width=4.4cm]{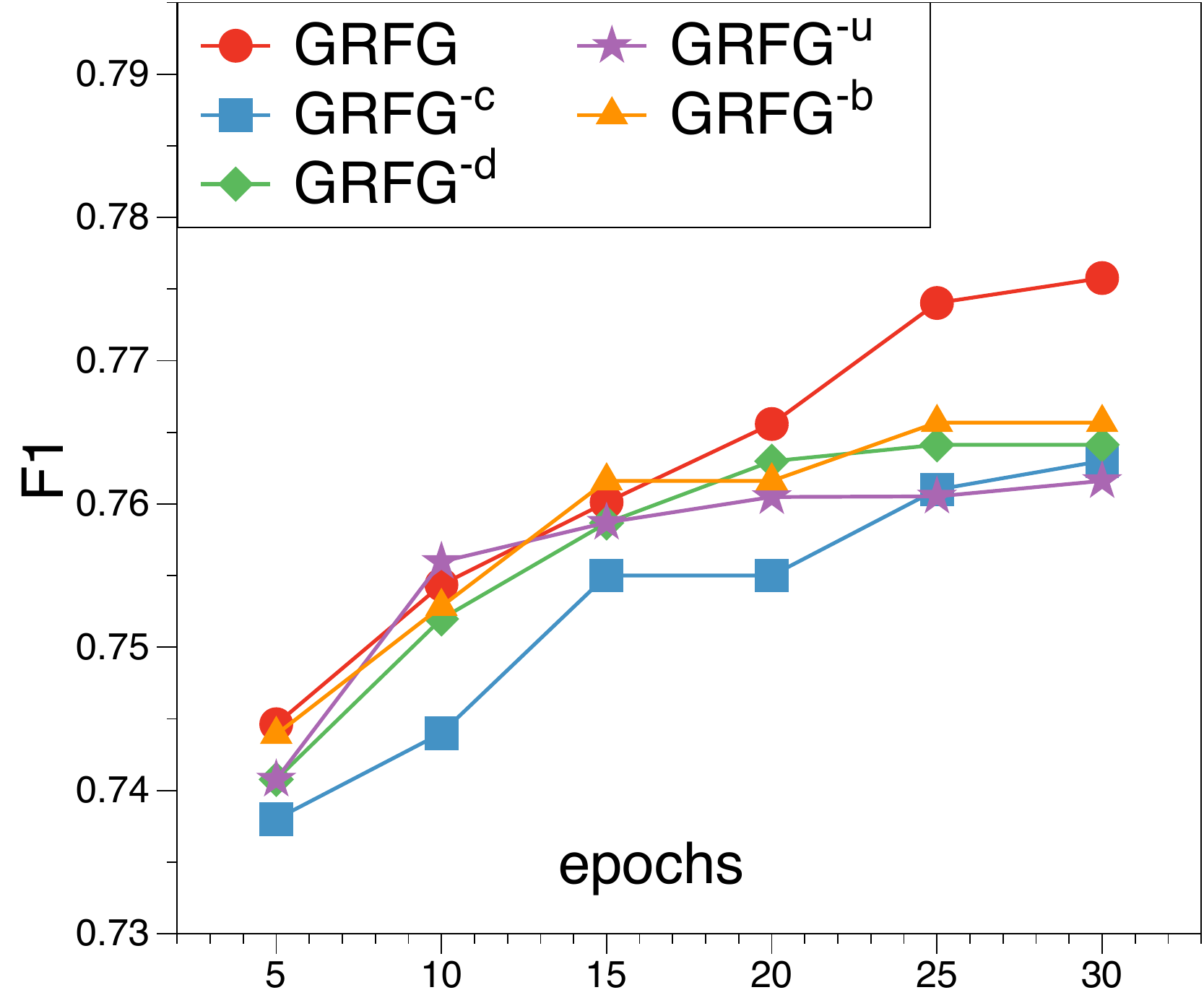}
}
\hspace{-3mm}
\subfigure[German Credit]{ 
\includegraphics[width=4.4cm]{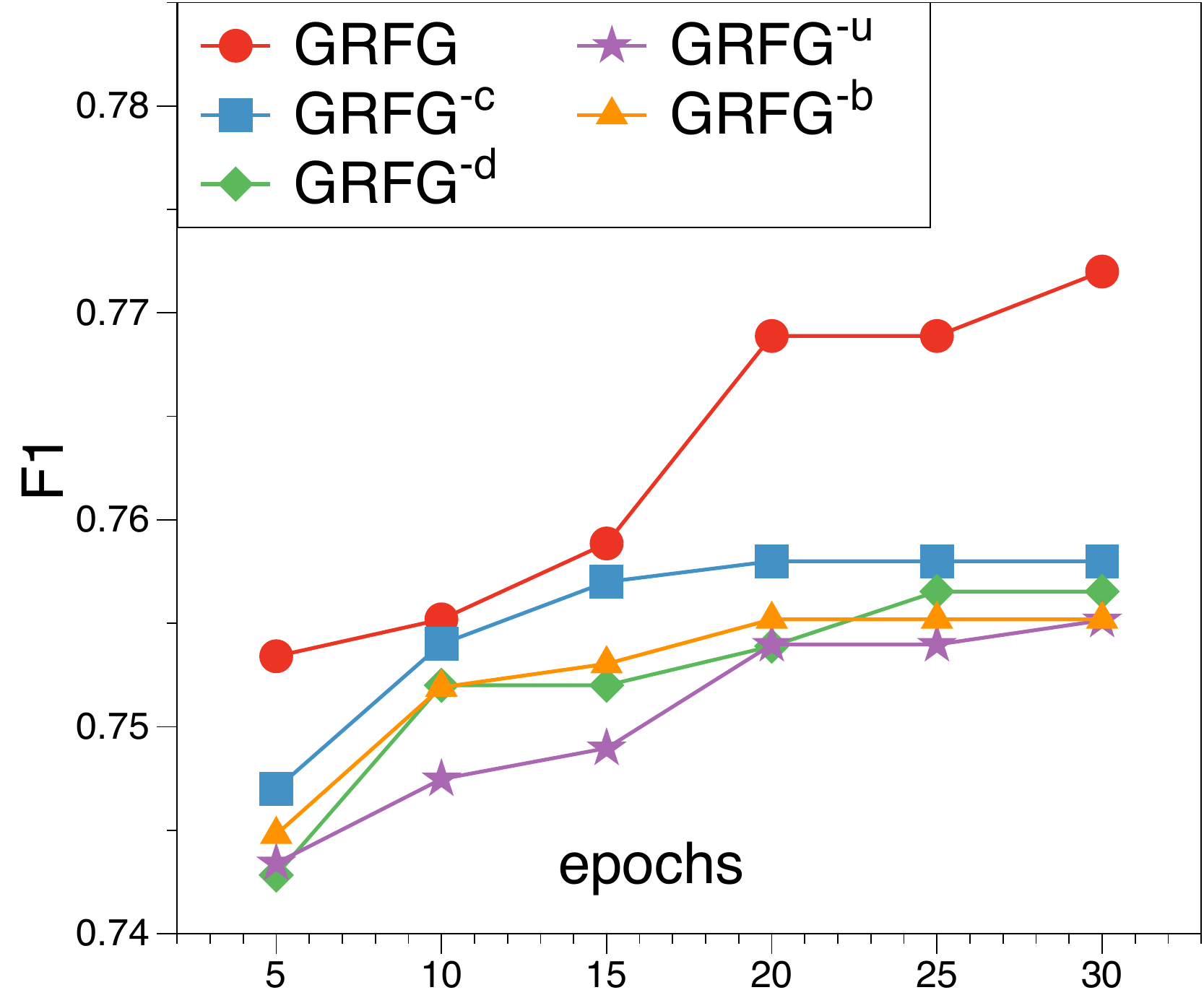}
}
\hspace{-3mm}
\subfigure[Housing Boston]{
\includegraphics[width=4.4cm]{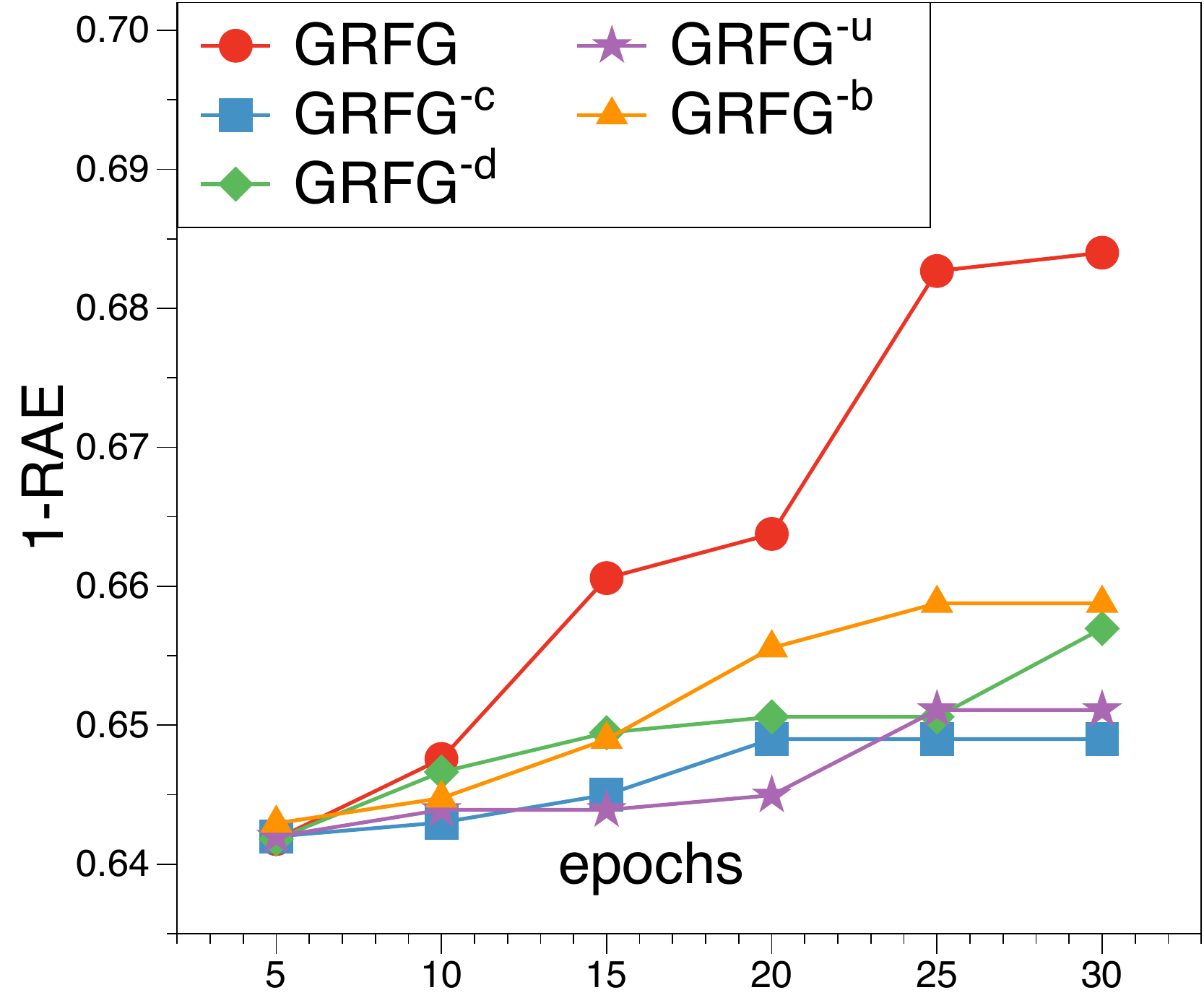}
}
\hspace{-3mm}
\subfigure[Openml\_589]{ 
\includegraphics[width=4.4cm]{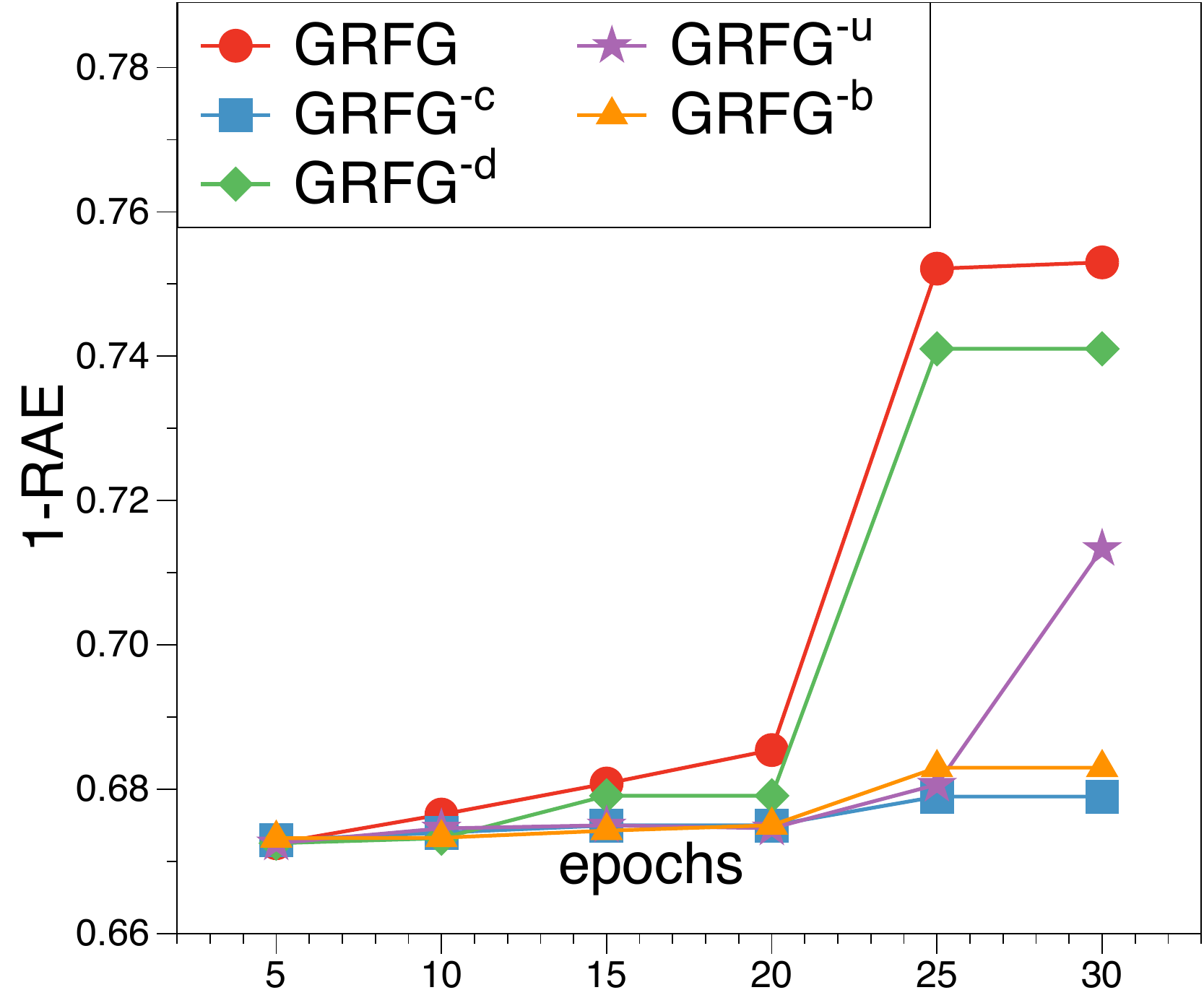}
}
\vspace{-0.35cm}
\caption{Comparison of different GRFG variants in terms of F1 or 1-RAE.}
\label{ab_study}
\vspace{-0.45cm}
\end{figure*}

\subsubsection{Study of the impact of each technical component}
\label{study_lp}
This experiment aims to answer:
\textit{How does each component in GRFG impact its performance?}
We developed four variants of GRFG (Section \ref{baseline}). 
Figure ~\ref{ab_study} shows the comparison results in terms of F1 score or 1-RAE on two classification datasests (\textit{i.e.} PimaIndian, German Credit) and two regression datasets (\textit{i.e.} Housing Boston, Openml\_589). 
First, we developed GRFG$^{-c}$ by removing the feature clustering step of GRFG. But, GRFG$^{-c}$ performs poorer than GRFG on all datasets. This shows the idea of group-level generation can augment reward feedback to help cascading agents learn better policies. 
Second, we developed  GRFG$^{-d}$ by using euclidean distance as feature distance metric in the M-clustering of GRFG.
The superiority of GRFG over GRFG$^{-d}$ suggests that our distance describes group-level information  relevance and redundancy ratio in order to maximize information distinctness across feature groups and minimize it within a feature group. Such a distance can help GRFG generate more useful dimensions.
Third, we developed GRFG$^{-u}$ and GRFG$^{-b}$ by using random in the two feature generation scenarios (Section \ref{group_generation}) of  GRFG.
We observed that GRFG$^{-u}$ and GRFG$^{-b}$ perform poorer than GRFG. This validates that crossing two distinct features and relevance prioritized generation can generate better features.

\begin{figure*}[htbp]
\centering
\subfigure[PimaIndian]{
\includegraphics[width=4.4cm]{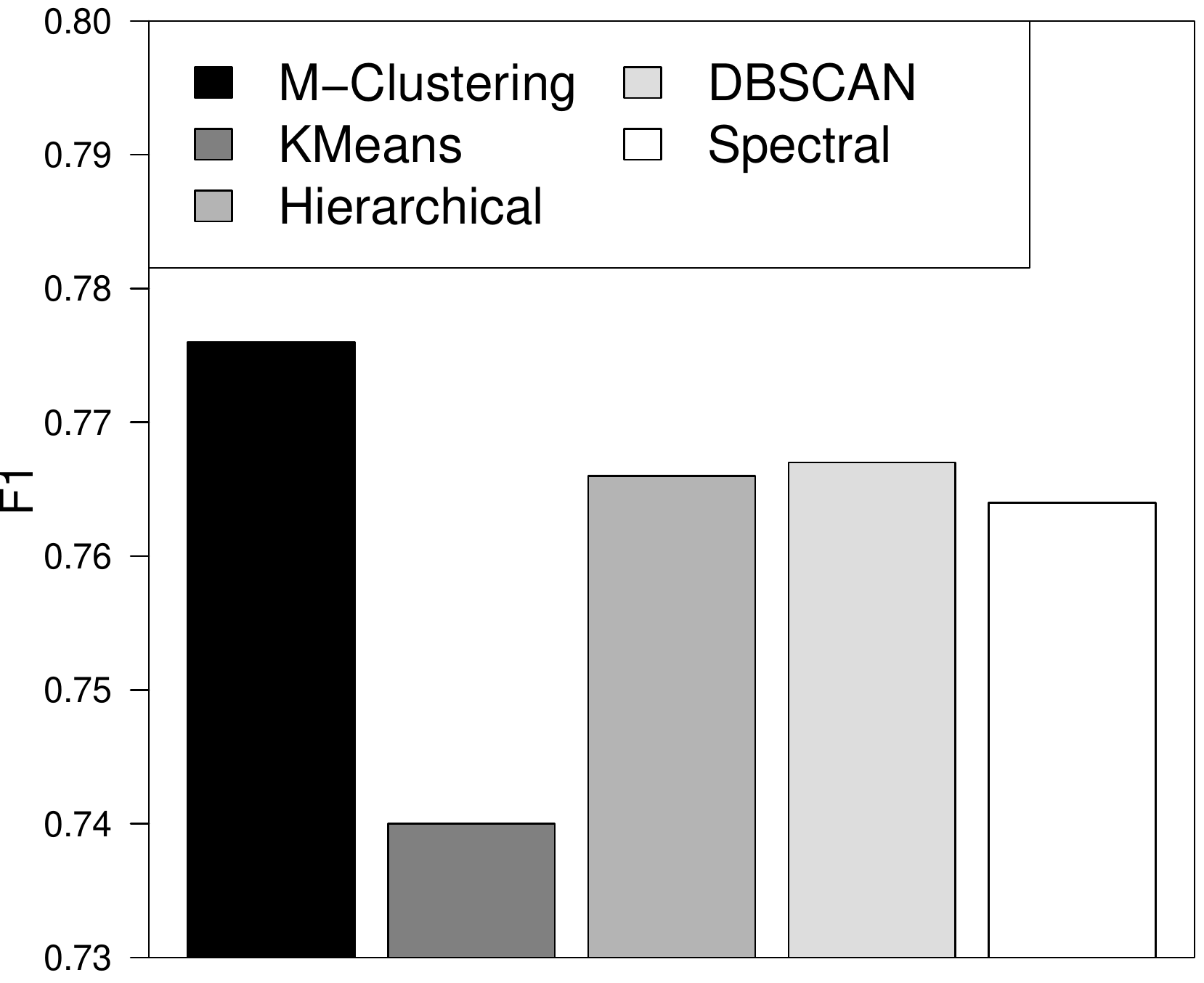}
}
\hspace{-3mm}
\subfigure[German Credit]{ 
\includegraphics[width=4.4cm]{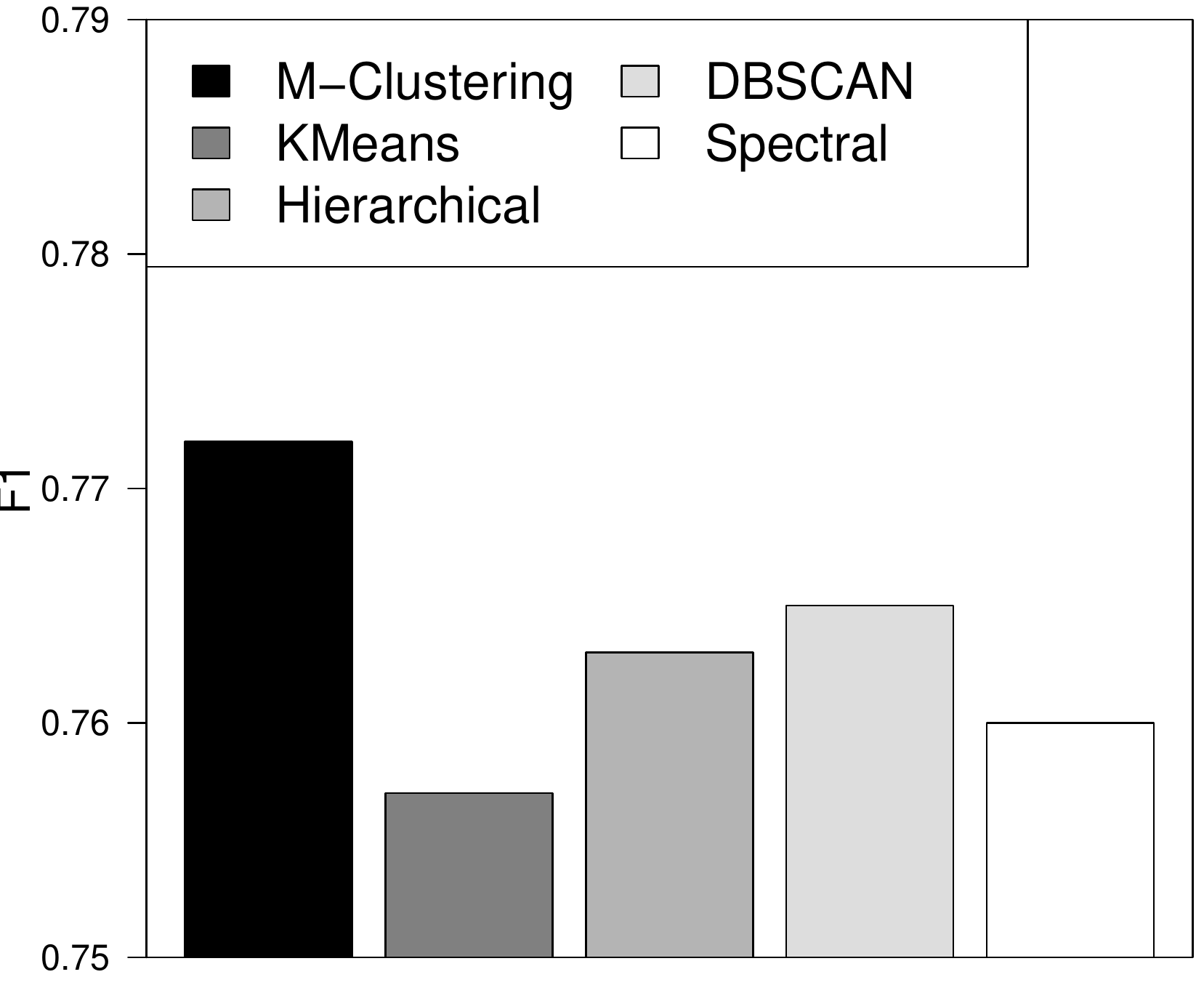}
}
\hspace{-3mm}
\subfigure[Housing Boston]{
\includegraphics[width=4.4cm]{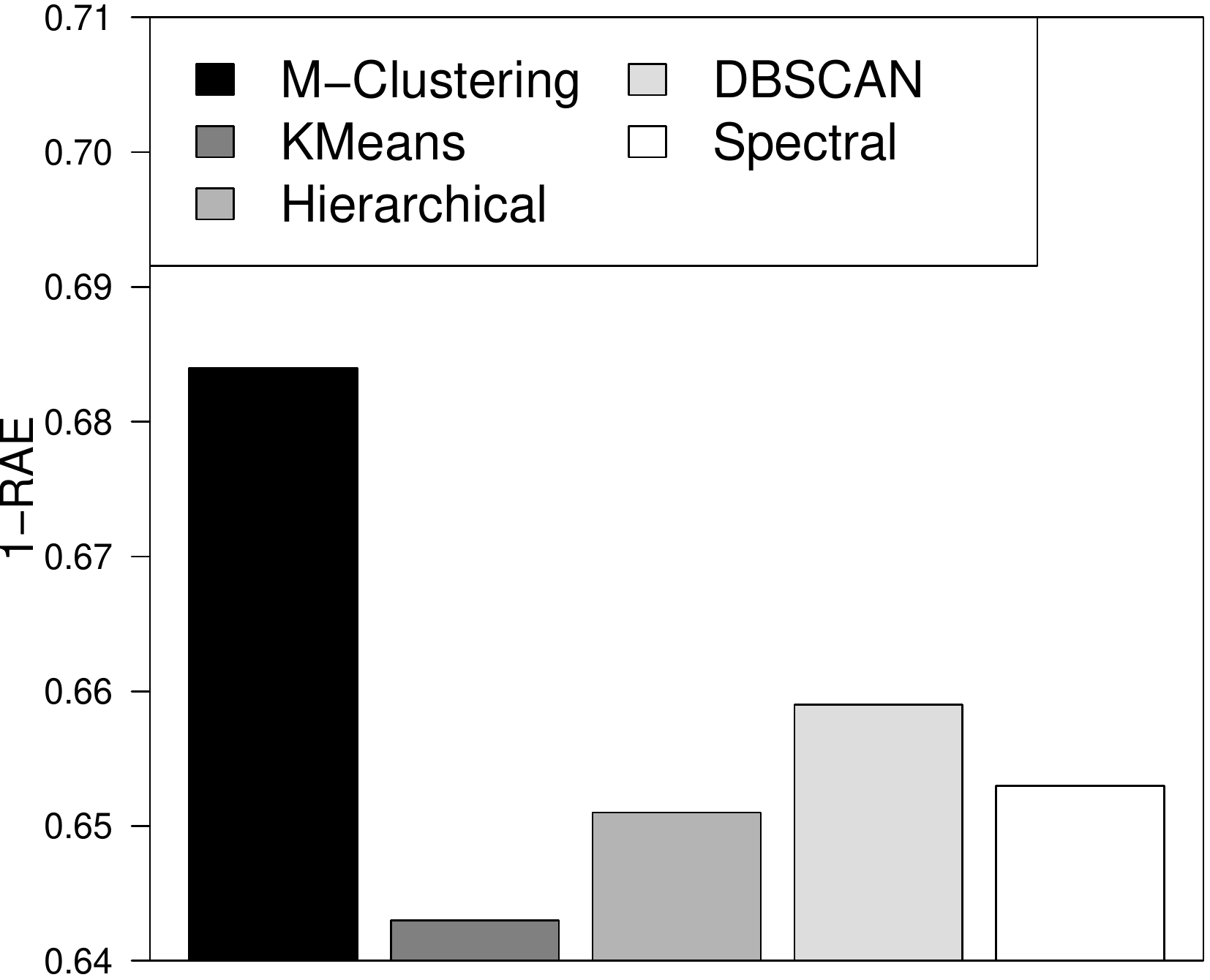}
}
\hspace{-3mm}
\subfigure[Openml\_589]{ 
\includegraphics[width=4.4cm]{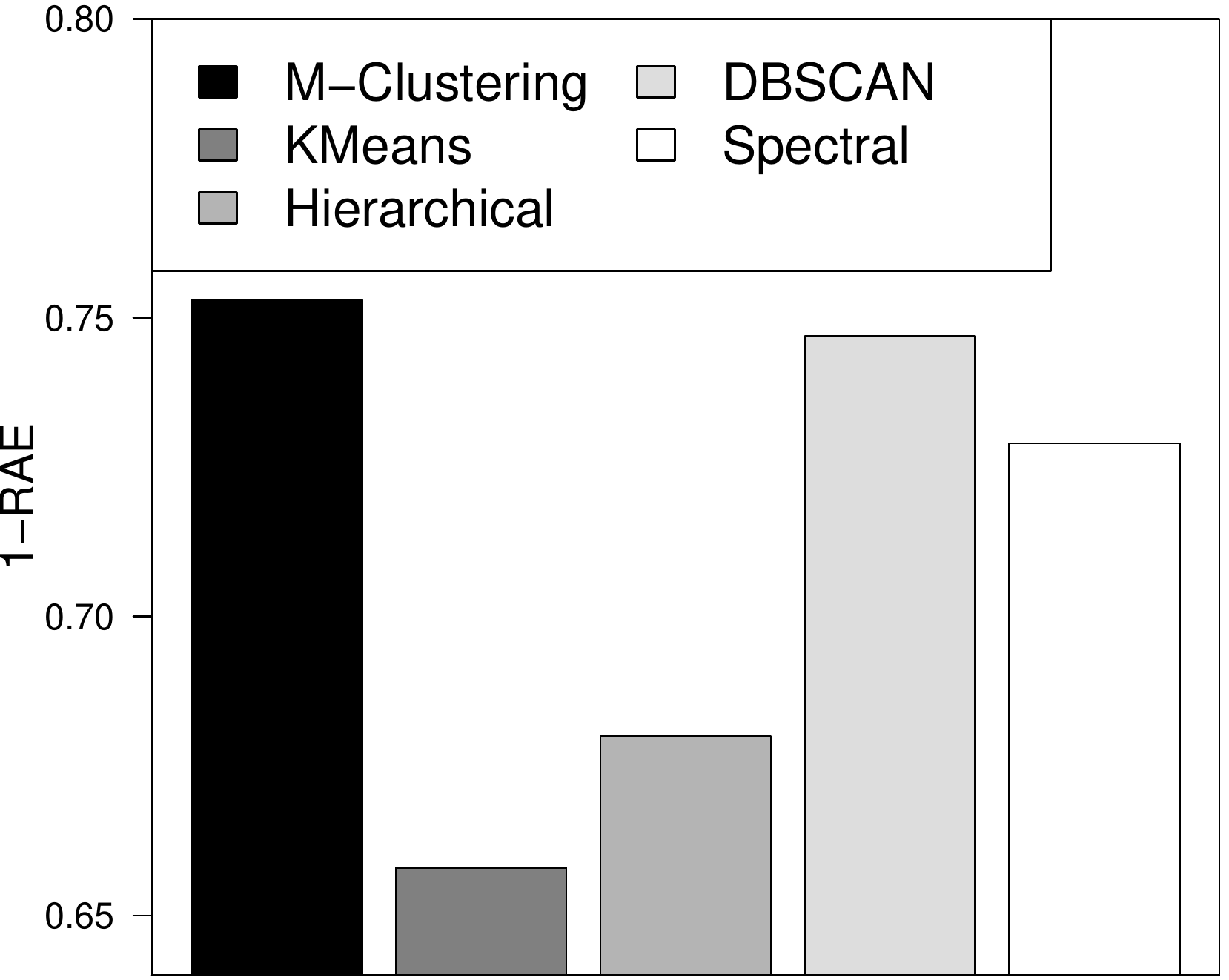}
}
\vspace{-0.35cm}
\caption{Comparison of different clustering algorithms in terms of F1 or 1-RAE.}
\label{differ_cluster}
\vspace{-0.cm}
\end{figure*}

\begin{figure*}[htbp]
\vspace{-0.15cm}
\centering
\subfigure[PimaIndian]{
\includegraphics[width=4.4cm]{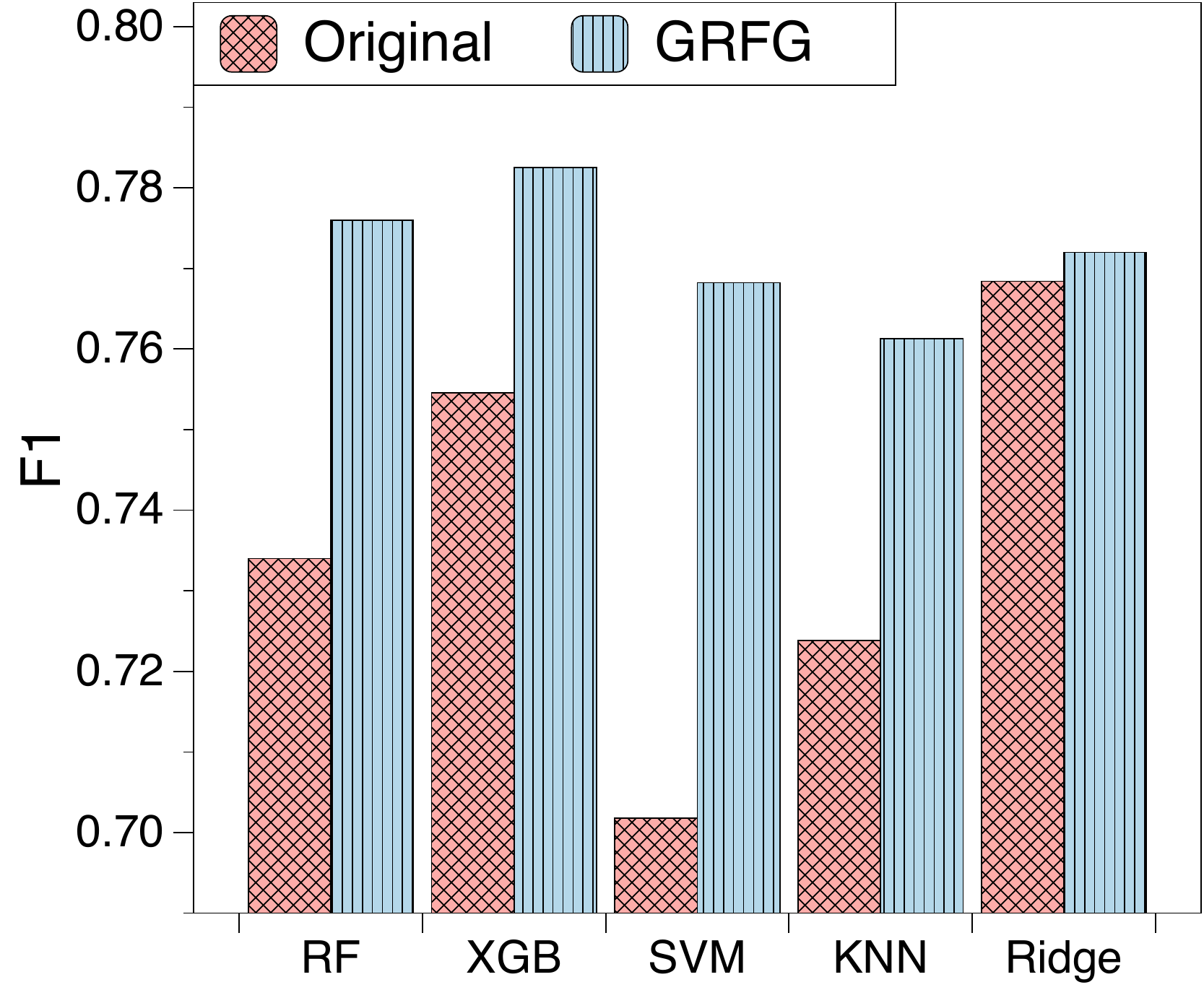}
}
\hspace{-3mm}
\subfigure[German Credit]{ 
\includegraphics[width=4.4cm]{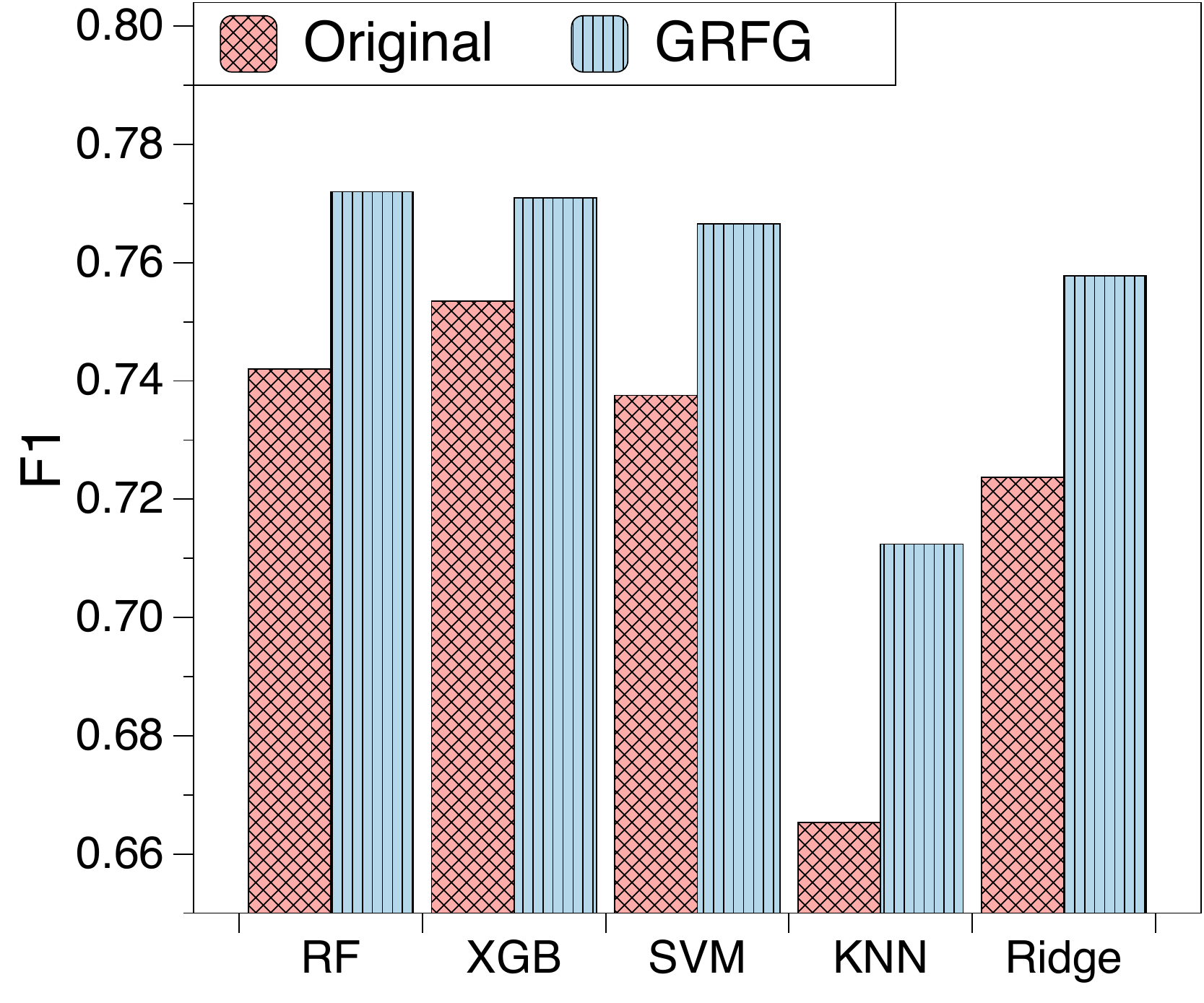}
}
\hspace{-3mm}
\subfigure[Housing Boston]{
\includegraphics[width=4.4cm]{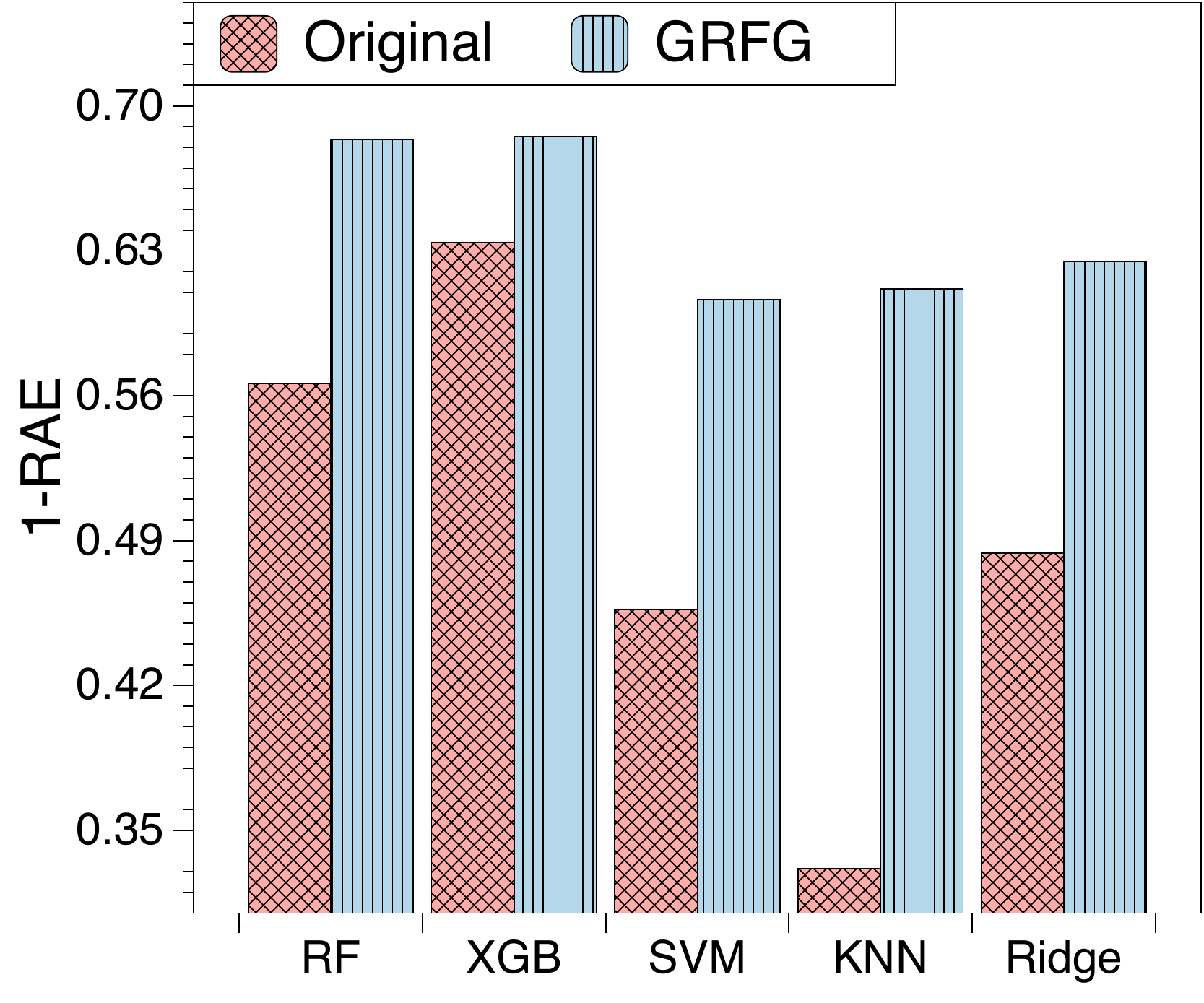}
}
\hspace{-3mm}
\subfigure[Openml 589]{ 
\includegraphics[width=4.4cm]{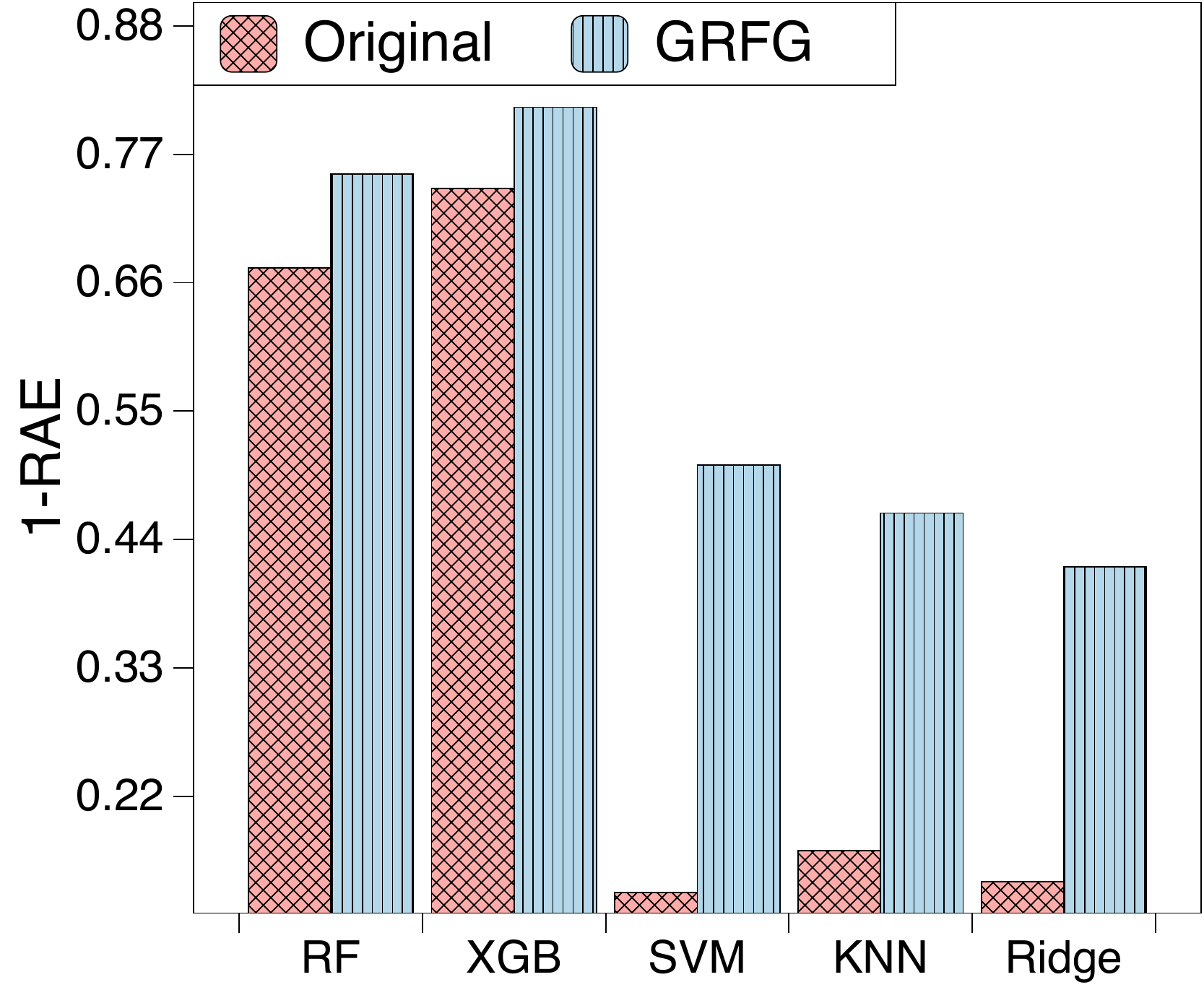}
}
\vspace{-0.35cm}
\caption{Comparison of different machine learning models in terms of F1 or 1-RAE.}
\label{differ_ml}
\vspace{-0.4cm}
\end{figure*}

\subsubsection{Study of the impact of M-Clustering}
This experiment aims to answer: \textit{Is M-Clustering more effective in improving feature generation than classical clustering algorithms?}
We replaced the feature clustering algorithm in GRFG with KMeans, Hierarchical Clustering, DBSCAN, and Spectral Clustering respectively.
We reported the comparison results in terms of F1 score or 1-RAE on the datasets used in Section \ref{study_lp}.
Figure \ref{differ_cluster} shows M-Clustering beats classical clustering algorithms on all datasets.
The underlying driver is that when feature sets change during generation, M-Clustering is more effective in minimizing information overlap of intra-group features and maximizing information distinctness of  inter-group features. So, crossing  the feature groups with distinct information is easier to generate  meaningful dimensions.

\subsubsection{Robustness check of GRFG under different machine learning (ML) models.}
This experiment is to answer:
\textit{Is GRFG robust when different ML models are used as a downstream task?}
We examined the robustness of GRFG by changing the ML model of a downstream task to Random Forest (RF), Xgboost (XGB), SVM, KNN, and Ridge Regression, respectively.
Figure ~\ref{differ_ml} shows the comparison results in terms of F1 score or 1-RAE on the datasets used in the Section \ref{study_lp}.
We observed that GRFG robustly improves model performances regardless of the ML model used.
This observation indicates that GRFG can generalize well to various benchmark applications and ML models.
We found that  RF and XGB are the two most powerful and robust predictors over the four datasets, which is consistent with the finding in Kaggle.COM competition community. 
Intuitively,  the accuracy of RF and XGB usually represent the performance ceiling on modeling a dataset. It is hard to break the performance ceiling.
But, after using our method to reconstruct the data, we continue to significantly improve the accuracy of RF and XGB and break through the performance ceiling.   
This finding clearly validates the strong robustness of our method.

\begin{figure}[!t]
\centering
\subfigure[Original Feature Space]{
\includegraphics[width=4.0cm]{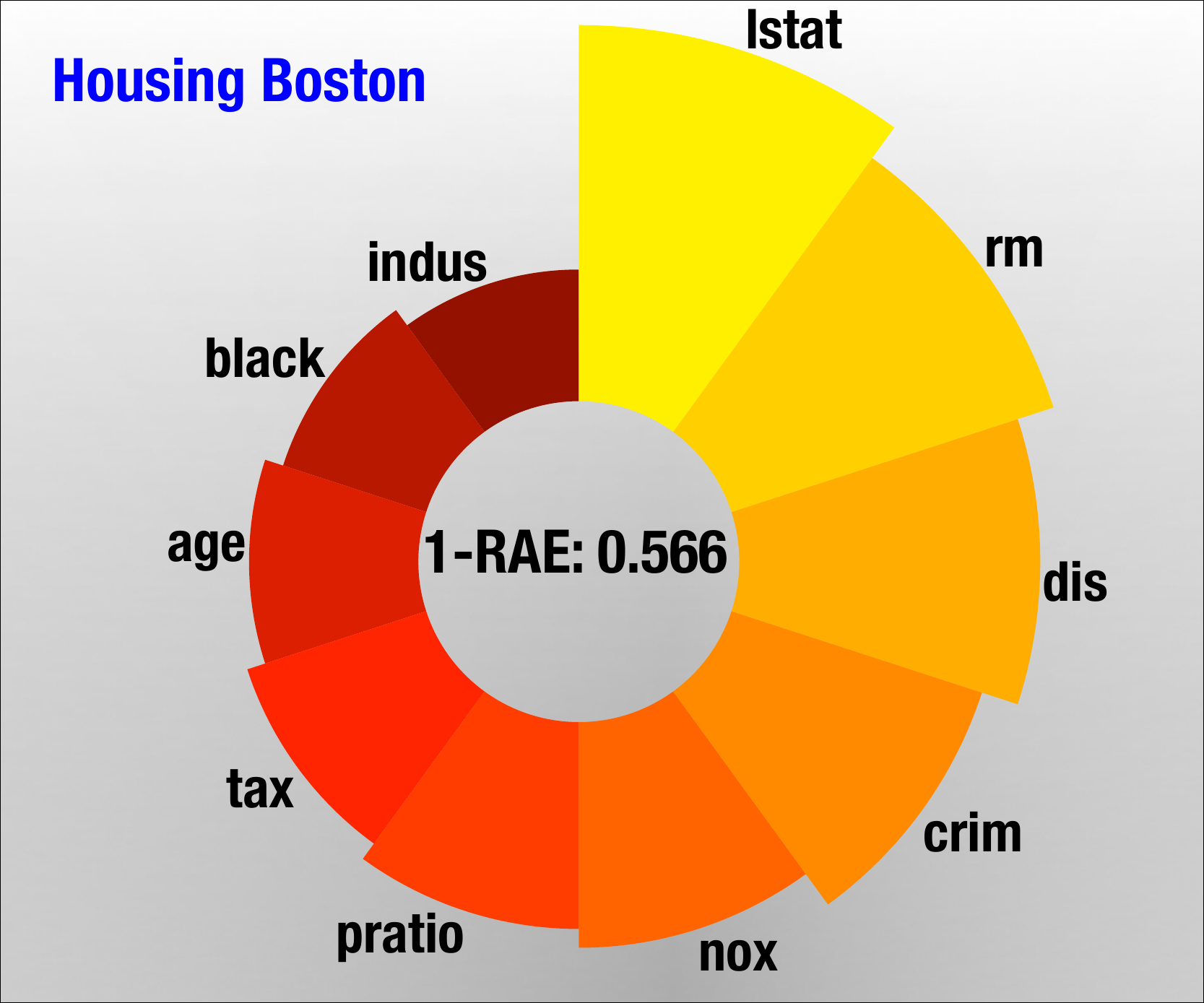}
}
\hspace{0mm}
\subfigure[GRFG-reconstructed Feature Space]{ 
\includegraphics[width=4.0cm]{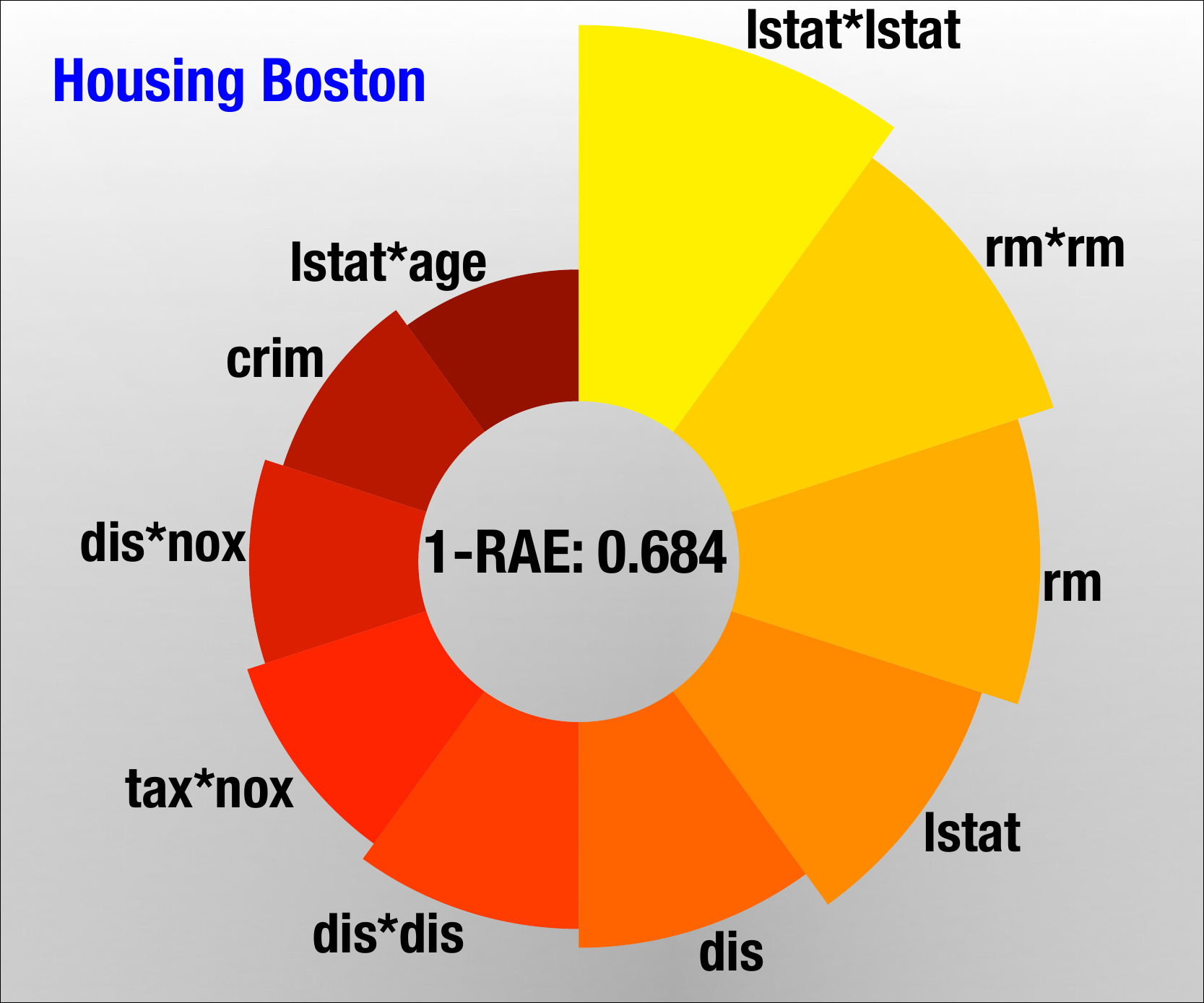}
}
\vspace{-0.4cm}
\caption{Top10 features for prediction in the original and GRFG-reconstructed feature space.}
\label{explain}
\vspace{-0.5cm}
\end{figure}


\subsubsection{Study of the traceability and explainability of GRFG}
This experiment aims to answer: \textit{Can GRFG generate an explainable feature space? Is this generation process traceable?}
We identified the top 10 essential features for prediction in both the original and reconstructed feature space using the Housing Boston dataset to predict housing prices with random forest regression.
Figure ~\ref{explain} shows the model performances in the central parts of each sub-figure. 
The texts associated with each pie chart describe the feature name.
If the feature name does not include an operation, the corresponding feature is original; otherwise, it is a generated feature. 
The larger the pie area is, the more essential the corresponding feature is.
We observed that the GRFG-reconstructed feature space greatly enhances the model performance by 20.9$\%$ and the generated features cover 60$\%$ of the top 10 features.
This indicates that GRFG generates informative features to refine the feature space.
Moreover, we can explicitly trace and explain the source and effect of a feature by checking its name.
For instance,  the feature ``lstat'' measures the percentage of the lower status populations in a house, which is negatively related to housing prices.
The most essential feature in the reconstructed feature space is 
``lstat*lstat'' that is generated by applying a ``multiply'' operation to ``lstat''.
This shows the generation process is traceable and the relationship between ``lstat'' and  housing prices is non-linear.

\section{Related Works}
\noindent\textbf{Reinforcement Learning (RL)} is the study of how intelligent agents should act in a given environment in order to maximize the expectation of cumulative rewards ~\cite{sutton2018reinforcement}.
According to the learned policy, we may classify reinforcement learning algorithms into two categories: value-based and policy-based.
Value-based algorithms (\textit{e.g.} DQN ~\cite{mnih2013playing}, Double DQN ~\cite{van2016deep}) estimate the value of the state or state-action pair for action selection.
Policy-based algorithms (\textit{e.g.} PG ~\cite{sutton2000policy}) learn a probability distribution to map state to action for action selection.
Additionally, an actor-critic reinforcement learning framework is proposed to  incorporate the advantages of value-based and policy-based algorithms ~\cite{schulman2017proximal}.
In recent years, RL has been applied to many domains (e.g. spatial-temporal data mining, recommended systems) and achieves great achievements ~\cite{wang2022reinforced,wang2022multi}.
In this paper, we formulate the selection of feature groups and operation as MDPs and propose a new cascading agent structure to resolve these MDPs.

\noindent\textbf{Automated Feature Engineering} aims to enhance the feature space through feature generation and feature selection in order to improve the performance of machine learning models ~\cite{chen2021techniques}.
Feature selection is to remove redundant features and retain important ones, whereas feature generation is to create and add meaningful variables. 
\ul{\textit{Feature Selection}} approaches include:
(i) filter methods (\textit{e.g}., univariate selection \cite{forman2003extensive}, correlation based selection \cite{yu2003feature}), in which features are ranked by a specific score like redundancy, relevance;  (ii) wrapper methods (\textit{e.g.}, Reinforcement Learning ~\cite{ liu2021efficient}, Branch and Bound~\cite{ kohavi1997wrappers}), in which the optimized feature subset is identified by a search strategy under a predictive task;  (iii) embedded methods (\textit{e.g.}, LASSO \cite{tibshirani1996regression}, decision tree \cite{sugumaran2007feature}), in which selection is part of the optimization objective of a predictive task. 
\ul{\textit{Feature Generation}} methods include: (i) latent representation learning based methods, e.g. latent dirichlet allocation ~\cite{blei2003latent}, deep factorization machine ~\cite{guo2017deepfm}, deep representation learning ~\cite{bengio2013representation}. 
Due to the latent feature space generated by these methods, it is hard to trace and explain the feature extraction process.
(ii) feature transformation based methods, which use arithmetic or aggregate operations to generate new features ~\cite{khurana2018feature,chen2019neural}.
These approaches have two weaknesses: (a) ignore feature-feature heterogeneity  among different feature pairs; (b) grow exponentially when the number of exploration steps increases.
Compared with prior literature, our personalized feature crossing strategy captures the feature distinctness,  cascading agents learn effective feature interaction policies, and  group-wise generation manner accelerates feature generation.

\section{Conclusion Remarks}
We present a group-wise reinforcement feature generation (GRFG) framework for optimal and explainable representation space reconstruction to  improve the performances of predictive models. 
This framework nests feature generation and selection in order to iteratively reconstruct a recognizable and size-controllable feature space via feature-crossing.
Specifically, first, we decompose the process of selecting crossing features and operations into three MDPs and develop a new cascading agent structure for it.
Second, we provide two feature generation strategies based on cosine similarity and mutual information  to deal with two generation scenarios following cascading selection.
Third, we suggest a group-wise feature generation manner to 
efficiently generate features and augment the rewards of cascading agents.
To accomplish this, we propose a new feature clustering algorithm (M-Clustering) to produce robust feature groups from an information theory perspective.
Through extensive experiments,
we can find that GRFG is effective at refining the feature space and shows competitive results compared to other baselines.
Moreover, GRFG can provide traceable  routes for feature generation, which improves the explainability of the refined feature space.
In the future, we aim to include the pre-training technique into GRFG  to further enhance feature generation.

\vspace{-0.12cm}

\bibliographystyle{ACM-Reference-Format}
\bibliography{acm,Yanjie}


\begin{thebibliography}{33}


\ifx \showCODEN    \undefined \def \showCODEN     #1{\unskip}     \fi
\ifx \showDOI      \undefined \def \showDOI       #1{#1}\fi
\ifx \showISBNx    \undefined \def \showISBNx     #1{\unskip}     \fi
\ifx \showISBNxiii \undefined \def \showISBNxiii  #1{\unskip}     \fi
\ifx \showISSN     \undefined \def \showISSN      #1{\unskip}     \fi
\ifx \showLCCN     \undefined \def \showLCCN      #1{\unskip}     \fi
\ifx \shownote     \undefined \def \shownote      #1{#1}          \fi
\ifx \showarticletitle \undefined \def \showarticletitle #1{#1}   \fi
\ifx \showURL      \undefined \def \showURL       {\relax}        \fi
\providecommand\bibfield[2]{#2}
\providecommand\bibinfo[2]{#2}
\providecommand\natexlab[1]{#1}
\providecommand\showeprint[2][]{arXiv:#2}

\bibitem[Bengio et~al\mbox{.}(2013)]%
        {bengio2013representation}
\bibfield{author}{\bibinfo{person}{Yoshua Bengio}, \bibinfo{person}{Aaron
  Courville}, {and} \bibinfo{person}{Pascal Vincent}.}
  \bibinfo{year}{2013}\natexlab{}.
\newblock \showarticletitle{Representation learning: A review and new
  perspectives}.
\newblock \bibinfo{journal}{\emph{IEEE transactions on pattern analysis and
  machine intelligence}} \bibinfo{volume}{35}, \bibinfo{number}{8}
  (\bibinfo{year}{2013}), \bibinfo{pages}{1798--1828}.
\newblock


\bibitem[Blei et~al\mbox{.}(2003)]%
        {blei2003latent}
\bibfield{author}{\bibinfo{person}{David~M Blei}, \bibinfo{person}{Andrew~Y
  Ng}, {and} \bibinfo{person}{Michael~I Jordan}.}
  \bibinfo{year}{2003}\natexlab{}.
\newblock \showarticletitle{Latent dirichlet allocation}.
\newblock \bibinfo{journal}{\emph{the Journal of machine Learning research}}
  \bibinfo{volume}{3} (\bibinfo{year}{2003}), \bibinfo{pages}{993--1022}.
\newblock


\bibitem[Cand{\`e}s et~al\mbox{.}(2011)]%
        {candes2011robust}
\bibfield{author}{\bibinfo{person}{Emmanuel~J Cand{\`e}s},
  \bibinfo{person}{Xiaodong Li}, \bibinfo{person}{Yi Ma}, {and}
  \bibinfo{person}{John Wright}.} \bibinfo{year}{2011}\natexlab{}.
\newblock \showarticletitle{Robust principal component analysis?}
\newblock \bibinfo{journal}{\emph{Journal of the ACM (JACM)}}
  \bibinfo{volume}{58}, \bibinfo{number}{3} (\bibinfo{year}{2011}),
  \bibinfo{pages}{1--37}.
\newblock


\bibitem[Chen et~al\mbox{.}(2019)]%
        {chen2019neural}
\bibfield{author}{\bibinfo{person}{Xiangning Chen}, \bibinfo{person}{Qingwei
  Lin}, \bibinfo{person}{Chuan Luo}, \bibinfo{person}{Xudong Li},
  \bibinfo{person}{Hongyu Zhang}, \bibinfo{person}{Yong Xu},
  \bibinfo{person}{Yingnong Dang}, \bibinfo{person}{Kaixin Sui},
  \bibinfo{person}{Xu Zhang}, \bibinfo{person}{Bo Qiao}, {et~al\mbox{.}}}
  \bibinfo{year}{2019}\natexlab{}.
\newblock \showarticletitle{Neural feature search: A neural architecture for
  automated feature engineering}. In \bibinfo{booktitle}{\emph{2019 IEEE
  International Conference on Data Mining (ICDM)}}. IEEE,
  \bibinfo{pages}{71--80}.
\newblock


\bibitem[Chen et~al\mbox{.}(2021)]%
        {chen2021techniques}
\bibfield{author}{\bibinfo{person}{Yi-Wei Chen}, \bibinfo{person}{Qingquan
  Song}, {and} \bibinfo{person}{Xia Hu}.} \bibinfo{year}{2021}\natexlab{}.
\newblock \showarticletitle{Techniques for automated machine learning}.
\newblock \bibinfo{journal}{\emph{ACM SIGKDD Explorations Newsletter}}
  \bibinfo{volume}{22}, \bibinfo{number}{2} (\bibinfo{year}{2021}),
  \bibinfo{pages}{35--50}.
\newblock


\bibitem[Chih-Jen(2022)]%
        {libsvm}
\bibfield{author}{\bibinfo{person}{Lin Chih-Jen}.}
  \bibinfo{year}{2022}\natexlab{}.
\newblock \bibinfo{title}{LibSVM Dataset Download}.
\newblock \bibinfo{howpublished}{[EB/OL]}.
\newblock
\newblock
\shownote{\url{https://www.csie.ntu.edu.tw/~cjlin/libsvmtools/datasets/}}.


\bibitem[Forman et~al\mbox{.}(2003)]%
        {forman2003extensive}
\bibfield{author}{\bibinfo{person}{George Forman} {et~al\mbox{.}}}
  \bibinfo{year}{2003}\natexlab{}.
\newblock \showarticletitle{An extensive empirical study of feature selection
  metrics for text classification.}
\newblock \bibinfo{journal}{\emph{J. Mach. Learn. Res.}} \bibinfo{volume}{3},
  \bibinfo{number}{Mar} (\bibinfo{year}{2003}), \bibinfo{pages}{1289--1305}.
\newblock


\bibitem[Friedman(2012)]%
        {friedman2012fast}
\bibfield{author}{\bibinfo{person}{Jerome~H Friedman}.}
  \bibinfo{year}{2012}\natexlab{}.
\newblock \showarticletitle{Fast sparse regression and classification}.
\newblock \bibinfo{journal}{\emph{International Journal of Forecasting}}
  \bibinfo{volume}{28}, \bibinfo{number}{3} (\bibinfo{year}{2012}),
  \bibinfo{pages}{722--738}.
\newblock


\bibitem[Fusi et~al\mbox{.}(2018)]%
        {fusi2018probabilistic}
\bibfield{author}{\bibinfo{person}{Nicolo Fusi}, \bibinfo{person}{Rishit
  Sheth}, {and} \bibinfo{person}{Melih Elibol}.}
  \bibinfo{year}{2018}\natexlab{}.
\newblock \showarticletitle{Probabilistic matrix factorization for automated
  machine learning}.
\newblock \bibinfo{journal}{\emph{Advances in neural information processing
  systems}}  \bibinfo{volume}{31} (\bibinfo{year}{2018}),
  \bibinfo{pages}{3348--3357}.
\newblock


\bibitem[Goyal and Ferrara(2018)]%
        {goyal2018graph}
\bibfield{author}{\bibinfo{person}{Palash Goyal} {and} \bibinfo{person}{Emilio
  Ferrara}.} \bibinfo{year}{2018}\natexlab{}.
\newblock \showarticletitle{Graph embedding techniques, applications, and
  performance: A survey}.
\newblock \bibinfo{journal}{\emph{Knowledge-Based Systems}}
  \bibinfo{volume}{151} (\bibinfo{year}{2018}), \bibinfo{pages}{78--94}.
\newblock


\bibitem[Guo et~al\mbox{.}(2017)]%
        {guo2017deepfm}
\bibfield{author}{\bibinfo{person}{Huifeng Guo}, \bibinfo{person}{Ruiming
  Tang}, \bibinfo{person}{Yunming Ye}, \bibinfo{person}{Zhenguo Li}, {and}
  \bibinfo{person}{Xiuqiang He}.} \bibinfo{year}{2017}\natexlab{}.
\newblock \showarticletitle{DeepFM: a factorization-machine based neural
  network for CTR prediction}.
\newblock \bibinfo{journal}{\emph{arXiv preprint arXiv:1703.04247}}
  (\bibinfo{year}{2017}).
\newblock


\bibitem[Guyon and Elisseeff(2003)]%
        {guyon2003introduction}
\bibfield{author}{\bibinfo{person}{I. Guyon} {and} \bibinfo{person}{A.
  Elisseeff}.} \bibinfo{year}{2003}\natexlab{}.
\newblock \showarticletitle{{An introduction to variable and feature
  selection}}.
\newblock \bibinfo{journal}{\emph{The Journal of Machine Learning Research}}
  \bibinfo{volume}{3} (\bibinfo{year}{2003}), \bibinfo{pages}{1157--1182}.
\newblock
\showISSN{1532-4435}


\bibitem[Hastie et~al\mbox{.}(2019)]%
        {hastie2019statistical}
\bibfield{author}{\bibinfo{person}{Trevor Hastie}, \bibinfo{person}{Robert
  Tibshirani}, {and} \bibinfo{person}{Martin Wainwright}.}
  \bibinfo{year}{2019}\natexlab{}.
\newblock \bibinfo{booktitle}{\emph{Statistical learning with sparsity: the
  lasso and generalizations}}.
\newblock \bibinfo{publisher}{Chapman and Hall/CRC}.
\newblock


\bibitem[Horn et~al\mbox{.}(2019)]%
        {horn2019autofeat}
\bibfield{author}{\bibinfo{person}{Franziska Horn}, \bibinfo{person}{Robert
  Pack}, {and} \bibinfo{person}{Michael Rieger}.}
  \bibinfo{year}{2019}\natexlab{}.
\newblock \showarticletitle{The autofeat python library for automated feature
  engineering and selection}.
\newblock \bibinfo{journal}{\emph{arXiv preprint arXiv:1901.07329}}
  (\bibinfo{year}{2019}).
\newblock


\bibitem[Howard(2022)]%
        {kaggle}
\bibfield{author}{\bibinfo{person}{Jeremy Howard}.}
  \bibinfo{year}{2022}\natexlab{}.
\newblock \bibinfo{title}{Kaggle Dataset Download}.
\newblock \bibinfo{howpublished}{[EB/OL]}.
\newblock
\newblock
\shownote{\url{https://www.kaggle.com/datasets}}.


\bibitem[Khurana et~al\mbox{.}(2018)]%
        {khurana2018feature}
\bibfield{author}{\bibinfo{person}{Udayan Khurana}, \bibinfo{person}{Horst
  Samulowitz}, {and} \bibinfo{person}{Deepak Turaga}.}
  \bibinfo{year}{2018}\natexlab{}.
\newblock \showarticletitle{Feature engineering for predictive modeling using
  reinforcement learning}. In \bibinfo{booktitle}{\emph{Proceedings of the AAAI
  Conference on Artificial Intelligence}}, Vol.~\bibinfo{volume}{32}.
\newblock


\bibitem[Kohavi and John(1997)]%
        {kohavi1997wrappers}
\bibfield{author}{\bibinfo{person}{Ron Kohavi} {and} \bibinfo{person}{George~H
  John}.} \bibinfo{year}{1997}\natexlab{}.
\newblock \showarticletitle{Wrappers for feature subset selection}.
\newblock \bibinfo{journal}{\emph{Artificial intelligence}}
  \bibinfo{volume}{97}, \bibinfo{number}{1-2} (\bibinfo{year}{1997}),
  \bibinfo{pages}{273--324}.
\newblock


\bibitem[Li et~al\mbox{.}(2017)]%
        {li2017feature}
\bibfield{author}{\bibinfo{person}{Jundong Li}, \bibinfo{person}{Kewei Cheng},
  \bibinfo{person}{Suhang Wang}, \bibinfo{person}{Fred Morstatter},
  \bibinfo{person}{Robert~P Trevino}, \bibinfo{person}{Jiliang Tang}, {and}
  \bibinfo{person}{Huan Liu}.} \bibinfo{year}{2017}\natexlab{}.
\newblock \showarticletitle{Feature selection: A data perspective}.
\newblock \bibinfo{journal}{\emph{ACM Computing Surveys (CSUR)}}
  \bibinfo{volume}{50}, \bibinfo{number}{6} (\bibinfo{year}{2017}),
  \bibinfo{pages}{1--45}.
\newblock


\bibitem[Liu et~al\mbox{.}(2021)]%
        {liu2021efficient}
\bibfield{author}{\bibinfo{person}{Kunpeng Liu}, \bibinfo{person}{Pengfei
  Wang}, \bibinfo{person}{Dongjie Wang}, \bibinfo{person}{Wan Du},
  \bibinfo{person}{Dapeng~Oliver Wu}, {and} \bibinfo{person}{Yanjie Fu}.}
  \bibinfo{year}{2021}\natexlab{}.
\newblock \showarticletitle{Efficient Reinforced Feature Selection via Early
  Stopping Traverse Strategy}. In \bibinfo{booktitle}{\emph{2021 IEEE
  International Conference on Data Mining (ICDM)}}. IEEE,
  \bibinfo{pages}{399--408}.
\newblock


\bibitem[Mnih et~al\mbox{.}(2013)]%
        {mnih2013playing}
\bibfield{author}{\bibinfo{person}{Volodymyr Mnih}, \bibinfo{person}{Koray
  Kavukcuoglu}, \bibinfo{person}{David Silver}, \bibinfo{person}{Alex Graves},
  \bibinfo{person}{Ioannis Antonoglou}, \bibinfo{person}{Daan Wierstra}, {and}
  \bibinfo{person}{Martin Riedmiller}.} \bibinfo{year}{2013}\natexlab{}.
\newblock \showarticletitle{Playing atari with deep reinforcement learning}.
\newblock \bibinfo{journal}{\emph{arXiv preprint arXiv:1312.5602}}
  (\bibinfo{year}{2013}).
\newblock


\bibitem[Public(2022a)]%
        {openml}
\bibfield{author}{\bibinfo{person}{Public}.} \bibinfo{year}{2022}\natexlab{a}.
\newblock \bibinfo{title}{Openml Dataset Download}.
\newblock \bibinfo{howpublished}{[EB/OL]}.
\newblock
\newblock
\shownote{\url{https://www.openml.org}}.


\bibitem[Public(2022b)]%
        {uci}
\bibfield{author}{\bibinfo{person}{Public}.} \bibinfo{year}{2022}\natexlab{b}.
\newblock \bibinfo{title}{UCI Dataset Download}.
\newblock \bibinfo{howpublished}{[EB/OL]}.
\newblock
\newblock
\shownote{\url{https://archive.ics.uci.edu/}}.


\bibitem[Schulman et~al\mbox{.}(2017)]%
        {schulman2017proximal}
\bibfield{author}{\bibinfo{person}{John Schulman}, \bibinfo{person}{Filip
  Wolski}, \bibinfo{person}{Prafulla Dhariwal}, \bibinfo{person}{Alec Radford},
  {and} \bibinfo{person}{Oleg Klimov}.} \bibinfo{year}{2017}\natexlab{}.
\newblock \showarticletitle{Proximal policy optimization algorithms}.
\newblock \bibinfo{journal}{\emph{arXiv preprint arXiv:1707.06347}}
  (\bibinfo{year}{2017}).
\newblock


\bibitem[Sugumaran et~al\mbox{.}(2007)]%
        {sugumaran2007feature}
\bibfield{author}{\bibinfo{person}{V Sugumaran}, \bibinfo{person}{V
  Muralidharan}, {and} \bibinfo{person}{KI Ramachandran}.}
  \bibinfo{year}{2007}\natexlab{}.
\newblock \showarticletitle{Feature selection using decision tree and
  classification through proximal support vector machine for fault diagnostics
  of roller bearing}.
\newblock \bibinfo{journal}{\emph{Mechanical systems and signal processing}}
  \bibinfo{volume}{21}, \bibinfo{number}{2} (\bibinfo{year}{2007}),
  \bibinfo{pages}{930--942}.
\newblock


\bibitem[Sutton and Barto(2018)]%
        {sutton2018reinforcement}
\bibfield{author}{\bibinfo{person}{Richard~S Sutton} {and}
  \bibinfo{person}{Andrew~G Barto}.} \bibinfo{year}{2018}\natexlab{}.
\newblock \bibinfo{booktitle}{\emph{Reinforcement learning: An introduction}}.
\newblock \bibinfo{publisher}{MIT press}.
\newblock


\bibitem[Sutton et~al\mbox{.}(2000)]%
        {sutton2000policy}
\bibfield{author}{\bibinfo{person}{Richard~S Sutton}, \bibinfo{person}{David~A
  McAllester}, \bibinfo{person}{Satinder~P Singh}, {and}
  \bibinfo{person}{Yishay Mansour}.} \bibinfo{year}{2000}\natexlab{}.
\newblock \showarticletitle{Policy gradient methods for reinforcement learning
  with function approximation}. In \bibinfo{booktitle}{\emph{Advances in neural
  information processing systems}}. \bibinfo{pages}{1057--1063}.
\newblock


\bibitem[Tibshirani(1996)]%
        {tibshirani1996regression}
\bibfield{author}{\bibinfo{person}{Robert Tibshirani}.}
  \bibinfo{year}{1996}\natexlab{}.
\newblock \showarticletitle{Regression shrinkage and selection via the lasso}.
\newblock \bibinfo{journal}{\emph{Journal of the Royal Statistical Society:
  Series B (Methodological)}} \bibinfo{volume}{58}, \bibinfo{number}{1}
  (\bibinfo{year}{1996}), \bibinfo{pages}{267--288}.
\newblock


\bibitem[Van~Hasselt et~al\mbox{.}(2016)]%
        {van2016deep}
\bibfield{author}{\bibinfo{person}{Hado Van~Hasselt}, \bibinfo{person}{Arthur
  Guez}, {and} \bibinfo{person}{David Silver}.}
  \bibinfo{year}{2016}\natexlab{}.
\newblock \showarticletitle{Deep reinforcement learning with double
  q-learning}. In \bibinfo{booktitle}{\emph{Proceedings of the AAAI conference
  on artificial intelligence}}, Vol.~\bibinfo{volume}{30}.
\newblock


\bibitem[Wang et~al\mbox{.}(2021a)]%
        {wang2021automated}
\bibfield{author}{\bibinfo{person}{Dongjie Wang}, \bibinfo{person}{Kunpeng
  Liu}, \bibinfo{person}{David Mohaisen}, \bibinfo{person}{Pengyang Wang},
  \bibinfo{person}{Chang-Tien Lu}, {and} \bibinfo{person}{Yanjie Fu}.}
  \bibinfo{year}{2021}\natexlab{a}.
\newblock \showarticletitle{Automated Feature-Topic Pairing: Aligning Semantic
  and Embedding Spaces in Spatial Representation Learning}. In
  \bibinfo{booktitle}{\emph{Proceedings of the 29th International Conference on
  Advances in Geographic Information Systems}}. \bibinfo{pages}{450--453}.
\newblock


\bibitem[Wang et~al\mbox{.}(2022b)]%
        {wang2022reinforced}
\bibfield{author}{\bibinfo{person}{Dongjie Wang}, \bibinfo{person}{Pengyang
  Wang}, \bibinfo{person}{Yanjie Fu}, \bibinfo{person}{Kunpeng Liu},
  \bibinfo{person}{Hui Xiong}, {and} \bibinfo{person}{Charles~E Hughes}.}
  \bibinfo{year}{2022}\natexlab{b}.
\newblock \showarticletitle{Reinforced Imitative Graph Learning for Mobile User
  Profiling}.
\newblock \bibinfo{journal}{\emph{arXiv preprint arXiv:2203.06550}}
  (\bibinfo{year}{2022}).
\newblock


\bibitem[Wang et~al\mbox{.}(2021b)]%
        {wang2021reinforced}
\bibfield{author}{\bibinfo{person}{Dongjie Wang}, \bibinfo{person}{Pengyang
  Wang}, \bibinfo{person}{Kunpeng Liu}, \bibinfo{person}{Yuanchun Zhou},
  \bibinfo{person}{Charles~E Hughes}, {and} \bibinfo{person}{Yanjie Fu}.}
  \bibinfo{year}{2021}\natexlab{b}.
\newblock \showarticletitle{Reinforced Imitative Graph Representation Learning
  for Mobile User Profiling: An Adversarial Training Perspective}. In
  \bibinfo{booktitle}{\emph{Proceedings of the AAAI Conference on Artificial
  Intelligence}}, Vol.~\bibinfo{volume}{35}. \bibinfo{pages}{4410--4417}.
\newblock


\bibitem[Wang et~al\mbox{.}(2022a)]%
        {wang2022multi}
\bibfield{author}{\bibinfo{person}{Xiting Wang}, \bibinfo{person}{Kunpeng Liu},
  \bibinfo{person}{Dongjie Wang}, \bibinfo{person}{Le Wu},
  \bibinfo{person}{Yanjie Fu}, {and} \bibinfo{person}{Xing Xie}.}
  \bibinfo{year}{2022}\natexlab{a}.
\newblock \showarticletitle{Multi-level Recommendation Reasoning over Knowledge
  Graphs with Reinforcement Learning}. In \bibinfo{booktitle}{\emph{Proceedings
  of the ACM Web Conference 2022}}. \bibinfo{pages}{2098--2108}.
\newblock


\bibitem[Yu and Liu(2003)]%
        {yu2003feature}
\bibfield{author}{\bibinfo{person}{Lei Yu} {and} \bibinfo{person}{Huan Liu}.}
  \bibinfo{year}{2003}\natexlab{}.
\newblock \showarticletitle{Feature selection for high-dimensional data: A fast
  correlation-based filter solution}. In \bibinfo{booktitle}{\emph{Proceedings
  of the 20th international conference on machine learning (ICML-03)}}.
  \bibinfo{pages}{856--863}.
\newblock


\end{thebibliography}


\end{document}